\documentclass[lettersize,journal]{IEEEtran}
\usepackage{amsmath,amsfonts}
\usepackage{algorithmic}
\usepackage{array}
\usepackage[caption=false,font=normalsize,labelfont=sf,textfont=sf]{subfig}
\usepackage{textcomp}
\usepackage{stfloats}
\usepackage{url}

\usepackage[colorlinks, linkcolor=red,  anchorcolor=blue, citecolor=blue]{hyperref}

\usepackage{verbatim}
\usepackage{graphicx}

\usepackage{multirow}
\usepackage{makecell}
\usepackage{amssymb}

\def\BibTeX{{\rm B\kern-.05em{\sc i\kern-.025em b}\kern-.08em
    T\kern-.1667em\lower.7ex\hbox{E}\kern-.125emX}}
\usepackage{balance}

\begin{document}
\title{Unpaired Image Captioning by Image-level Weakly-Supervised Visual Concept Recognition}
\author{\makecell{Peipei Zhu, Xiao Wang, Yong Luo, Zhenglong Sun, \\ Wei-Shi Zheng, Yaowei Wang, \textit{Member, IEEE}, and Changwen Chen, \textit{Fellow, IEEE}}
\thanks{Peipei Zhu and Zhenglong Sun are with the School of Science and Engineering, The Chinese University of Hong Kong (Shen Zhen), Shen Zhen 518172, China. The work was done when the first author had an internship at Peng Cheng Laboratory, Shenzhen, China.

Xiao Wang, Wei-Shi Zheng, and Yaowei Wang are with Peng Cheng Laboratory, Shen Zhen, China. Wei-Shi Zheng is also with the School of Computer Science and Engineering, SUN YAT-SEN UNIVERSITY, Guang Zhou, China.

Yong Luo is with the School of Computer Science, Wuhan University, Wuhan, China.

Changwen Chen is with the Department of Computing, The Hong Kong Polytechnic University (PolyU), HongKong, China.

Yaowei Wang and Changwen Chen are the first and second corresponding authors, respectively. E-mail: wangyw@pcl.ac.cn, changwen.chen@polyu.edu.hk, peipeizhu@link.cuhk.edu.cn, wangxiaocvpr@foxmail.com, wszheng@ieee.org, luoyong@whu.edu.cn, sunzhenglong@cuhk.edu.cn.
}}

%\markboth{Journal of \LaTeX\ Class Files,~Vol.~18, No.~9, September~2020}
%{How to Use the IEEEtran \LaTeX \ Templates}

\maketitle

\begin{abstract}
The goal of unpaired image captioning (UIC) is to describe images without using image-caption pairs in the training phase. Although challenging, we except the task can be accomplished by leveraging a training set of images aligned with visual concepts. Most existing studies use off-the-shelf algorithms to obtain the visual concepts because the Bounding Box (BBox) labels or relationship-triplet labels used for the training are expensive to acquire. In order to resolve the problem in expensive annotations, we propose a novel approach to achieve cost-effective UIC. Specifically, we adopt image-level labels for the optimization of the UIC model in a weakly-supervised manner. For each image, we assume that only the image-level labels (such as object categories and the relationships) are available without specific locations and numbers. The image-level labels are utilized to train a weakly-supervised object recognition model to extract object information (e.g., instance) in an image, and the extracted instances are adopted to infer the relationships among different objects based on an enhanced graph neural network (GNN). The proposed approach achieves comparable or even better performance compared with previous methods without the expensive cost of annotations. Furthermore, we design an unrecognized object (UnO) loss combined with a visual concept reward to improve the alignment of the inferred object and relationship information with the images. It can effectively alleviate the issue encountered by existing UIC models about generating sentences with nonexistent objects. To the best of our knowledge, this is the first attempt to solve the problem of Weakly-Supervised visual concept recognition for UIC (WS-UIC) based only on image-level labels. Extensive experiments have been carried out to demonstrate that the proposed WS-UIC model achieves inspiring results on the COCO dataset while significantly reducing the cost of labeling.

% Extensive experiments demonstrate that our proposed WS-UIC model achieves inspiring results on the COCO dataset (even outperforms many previous approaches that need much stronger supervision), and significantly reduces the cost of labeling.
\end{abstract}


\section{Introduction}

Image captioning \cite{vinyals2015show, feng2019unsupervised, zhang2020integrating} aims at describing the content and event of an image using a couple of words. We can also treat it as a mapping from image to corresponding natural language descriptions \cite{wu2020fine}. This task has been greatly promoted by the deep learning algorithms \cite{vinyals2015show}, which can learn from large-scale annotated image-sentence pairs \cite{vinyals2016show}. Image captioning has been widely used in many practical applications \cite{hossain2019comprehensive}, including human-robot interaction \cite{das2017visual, ling2017teaching}, automatic driving \cite{kim2018textual, wang2018look}, visual assistance for impaired people \cite{wu2017automatic, gurari2018vizwiz}, etc. The essential practice of state-of-the-art image captioning approaches follows the encoder-decoder paradigm \cite{ben2021unpaired}. Generally, an input image is first encoded into feature representation using Convolutional Neural Network (CNN). Then, the Recurrent Neural Network (RNN) is adopted to decode the features into multiple words one by one. This way, the natural description for the input image can be obtained.

\begin{figure}[!t]
    \centering
    \includegraphics[width=8.5cm, height=5.5cm]{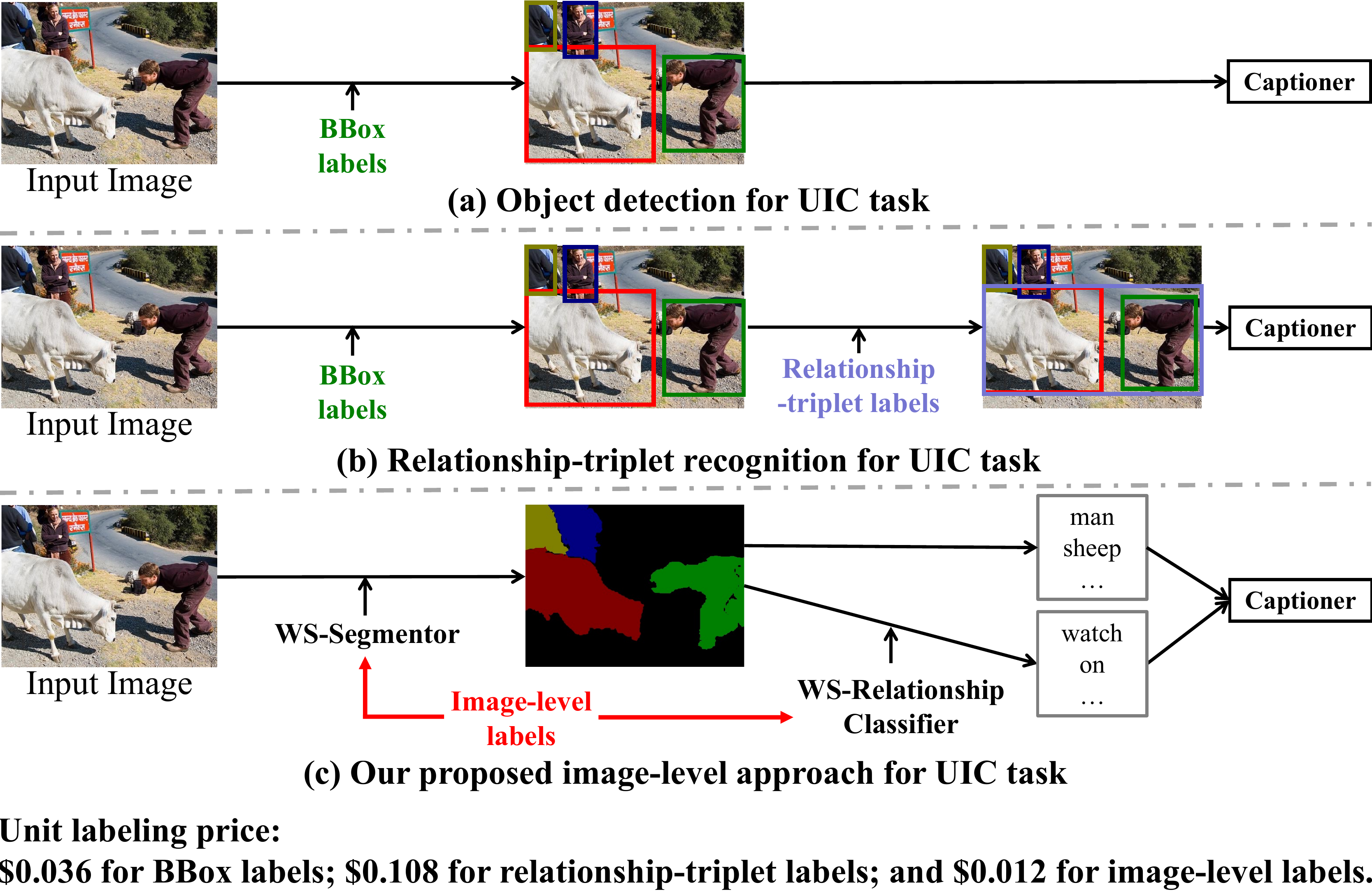}
    \caption{\textbf{Comparisons between the unpaired image captioning (UIC) approaches based on (a) \emph{object detection}, (b) \emph{relationship-triplet recognition}, and (c) the proposed weakly-supervised (WS) visual concept recognition that requires only image-level labels.} Specifically, the \textit{object detection} scheme utilizes BBox labels to recognize object concepts \cite{feng2019unsupervised, laina2019towards, guo2020recurrent}; the approach termed \textit{relationship-triplet recognition} combines \textit{object detection} and \textit{relationship-triplet recognition} \cite{gu2019unpaired, liu2019exploring, cao2020interactions} to differentiate the (subject, relationship, object) information, which relies on complicated relationship-triplet labels; and the proposed WS-UIC method, which is based on only image-level labels to train WS object recognition (WS-OR) and WS relationship recognition (WS-RR), can achieve better performance than most of the (a) and (b) approaches. According to the \textit{Amazon Mechanical Turk}, the unit price of labeling BBox labels and relationship labels are three times and nine times that of annotating image-level labels, respectively \cite{ipeirotis2010quality}.}
    \label{fig:pipeline}
\end{figure}

In spite of the promising applications and various mature models developed, the standard image captioning models are mostly trained in a fully-supervised manner, which may require a tremendous number of manually annotated image-caption pairs \cite{li2019know, yang2018multitask}. It is very difficult to obtain such image-caption pairs since the manual annotation is very costly \cite{mao2015learning, yao2017incorporating, hendricks2016deep}. In addition, the generalization ability of such models maybe limited, as the collected images and annotated captions are often biased and incomplete. The existing image captioning datasets, such as Microsoft COCO \cite{lin2014microsoft}, are relatively small in scales comparing with the most popular image classification datasets, such as OpenImages \cite{krasin2017openimages} and ImageNet \cite{deng2009imagenet}. The varieties of images and captions within these datasets are also limited in the order of 100 object categories \cite{feng2019unsupervised}. As a result, it is difficult for the captioning models trained on such paired image-caption data to generalize them to images in the wild. Therefore, it is desirable to develop Unpaired Image Captioning (UIC) approaches that do not require image-caption training pairs \cite{baldassarre2020explanation, liu2019exploring, liu2020bridging}.

In the absence of image-caption pairs, the learning of UIC models needs enormously additive labels. For example, the UIC model often needs to recognize the \emph{category} and \emph{attributes} of objects, and sometimes the \emph{relationships} between different objects in the image. The clues obtained can be utilized to build the connections between the visual concepts and the images for captioning \cite{hendricks2016deep}. Generally speaking, the visual concept recognition can be categorized into two streams, \textit{i.e.}, \textit{object detection} and \textit{relationship-triplet recognition}. As illustrated in Fig. \ref{fig:pipeline} (a) and (b), the \textit{object detection} based models rely on enormous and costly Bounding Box (BBox) labels to recognize the objects \cite{feng2019unsupervised, laina2019towards, guo2020recurrent}. The \textit{relationship-triplet recognition} based algorithms achieve better performance, but require enormous relationship-triplet $<$object, relationship, subject$>$ labels \cite{gu2019unpaired, liu2019exploring, cao2020interactions}. Although some off-the-shelf \textit{object detection} or \textit{relationship-triplet recognition} models can be adopted to achieve UIC, abundant BBox or relationship-triplet labels are required to be annotated when the UIC model is applied in a new scenario. This is because the new scenarios may contain quite different objects and new relationships of objects. Therefore, a natural question will be: \emph{\textbf{do we have any other cost-effective ways for the learning of the UIC models?}}

In this paper, we attempt to address the aforementioned problem by utilizing only the easily accessible \textbf{image-level labels}, which contain the statistical information (\textit{i.e.}, the category) of objects and relationships of the corresponding image. Comparing with existing expensive BBox or relationship-triplet labels based models \footnote{\url{https://aws.amazon.com/sagemaker/groundtruth/pricing/}}, the proposed algorithm can achieve comparable performance, depending only on inexpensive labels that are much more convenient to be annotated (\textit{i.e.}, much cheaper annotation cost). Such cost-effective annotations can be used for both object recognition and object relationship mining. As shown in Fig. \ref{fig:Whole_Arc}, given an image and the corresponding image-level labels, we train a weakly-supervised instance segmentation model inspired by \cite{ahn2019weakly}. Meanwhile, we exploit the multi-scale feature maps extracted from the backbone network and model the spatial relations of multiple-instance using graph neural network (GNN). To make the training of our model more stable and faster to converge, we also introduce BN (Batch Normalization) and the residual connections into GNN. Finally, the predicted object category information and the relationship information are utilized to guide the unpaired image captioning. Given an image, an encoder-decoder framework \cite{wu2019recall, guo2019show} is adopted for caption generation, where an unrecognized object (UnO) loss is designed to integrate with the concept reward loss for optimization. The UnO loss contributes to excluding the unrecognized objects in the caption generation phase and therefore addresses the issue of unknown target generation, which has impacted negatively on existing UIC models.

To sum up, the main contributions of this paper can be summarized into following three aspects: 

$\bullet$ We propose a novel weakly-supervised visual concept recognition framework, \textit{i.e.}, using the image-level labels only for UIC. This framework is capable of \textit{object and relationship} concept recognition using only image-level object labels and image-level relationship labels. The image-level labels are inexpensive to obtain, which reduces the annotation cost significantly. Moreover, such a framework can be applied to other weakly-supervised visual analytic tasks.

$\bullet$ We propose a novel Unrecognized Object (UnO) loss, which is the first attempt to wisely consider the unrecognized objects in UIC objectives, and address the unknown target object issue well.

$\bullet$ We propose an improved GNN-BR module to enhance the caption generation of the UIC model, where the batch normalization strategy can make the model more stable and easier to converge, and the residual block can alleviate the gradient degradation problem.

We also conduct extensive experiments on the popular COCO datasets \cite{lin2014microsoft}. The results demonstrate that this weakly-supervised approach with significantly reduced cost of labeling is still capable of achieving performance that is better than those approaches using much stronger supervision with millions of BBox labels.

\section{Related Work}
In this section, we first give a review of the image captioning and unpaired image captioning (UIC), which learns image captioner with paired image-sentence data or unpaired image/sentence data. Then, we discuss the core techniques for UIC, \textit{i.e.}, the object concept recognition and the relationship exploration. %More related works can be found in the following survey. 

\subsection{Image Captioning}
In the past few years, fully-supervised image captioning has been studied extensively \cite{hossain2019comprehensive}. The majority of the proposed models adopt the encoder-decoder paradigm where one Convolutional Neural Network (CNN) is leveraged to encode an input image firstly and one Recurrent Neural Network (RNN) is utilized to output a description for the image subsequently \cite{vinyals2015show, wu2019recall, xu2019multi}. These models are trained to maximize the probability of generating the ground-truth captions, depending on enormous image-caption pairs. As paired image-caption data is hard to collect, some researchers attempted to decouple the dependency on the paired annotations through other available datasets \cite{ben2021unpaired}. Hendricks \textit{et al.} \cite{hendricks2016deep} trained a caption model of describing novel objects without relying on image-caption data containing the novel object concepts, which leverages large object recognition datasets and external text corpora by transferring knowledge between semantically similar concepts. Yao \textit{et al.} \cite{yao2017incorporating} presented a Long Short-Term Memory with Copying Mechanism (LSTM-C) to describe novel objects in captions, incorporating copying mechanism into the CNN plus RNN image captioning framework. Chen et al. \cite{chen2016semi} proposed a semi-supervised image captioning model by artificially generating missing visual information conditioned on the textual data. Kim \textit{et al.} \cite{kim2019image} also proposed a semi-supervised learning method to assign pseudo-labels to unlabeled images via Generative Adversarial Networks, which in turn are utilized to train a fully-supervised captioner. Although promising captioning results have been achieved, the novel object captioning or semi-supervised image captioning methods still require expensive paired image-caption data for training. Different from these works, we aim to tackle unpaired image captioning without relying on any image-caption pair.

\subsection{Unpaired Image Captioning}
UIC is capable of producing captions for input images without adopting any paired image-sentence data and has attracted significant attention from researchers. Gu \textit{et al.} \cite{gu2018unpaired} implemented language pivoting to achieve unpaired image captioning. However, the scheme requires the ground-truth pivot-image pairs and paired pivot-target language translation datasets. Feng \textit{et al.} \cite{feng2019unsupervised} proposed the first work that tackles the captioning task through training with totally unpaired image-sentence datasets. Recently, Laina \textit{et al.} \cite{laina2019towards} employed a shared multi-modal embedding, structured by visual concepts, to bridge the gap between the image and sentence domains. Guo \textit{et al.} \cite{guo2020recurrent} also proposed a novel Recurrent Relational Memory (R$^2$M) Network which can be implemented to get rid of the complicated and sensitive adversarial learning. SCS \cite{ben2021unpaired} has proposed a semantic-constrained self-learning strategy that iteratively generates ``pseudo" sentences and re-trains the captioner for UIC. In recent years, other researchers have adopted scene graph modeling in the model to utilize more semantic information, including the relationships between objects and the attributes of objects \cite{gu2019unpaired, liu2019exploring, cao2020interactions}. All of these approaches achieved better performance than the schemes proposed by Feng \textit{et al.} and Laina \textit{et al.} but required much more ground-truth information or more expensive annotations.

Although an unpaired image-sentence dataset is used, these approaches still depend on enormously expensive labels at the visual concept recognition stage, including the BBox labels or the relationship-triplet labels. As far as we know, the proposed approach in this research is the first work to tackle visual concept recognition for UIC aided by only image-level class labels. Besides, an unrecognized object loss is designed to guide the UIC model to exclude the unrecognized objects.

\subsection{Captioning based on Object Concept Recognition}
The schemes of object concept recognition have been developed to recognize the objects contained in an image. The recognized object concepts ensure that the generated captions can represent the main topic of an image by aligning these object words with images. Various schemes have been developed, including image classification, object detection \cite{huang2017speed, ren2016faster, singh2018analysis}, Multiple Instance Learning (MIL), and so on. In particular, Hendricks \textit{et al.} \cite{hendricks2016deep} and Venugopalan \textit{et al.} \cite{venugopalan2017captioning} utilized an image classifier to recognize various object categories. However, this approach can only be implemented to differentiate the object categories instead of the instances. Other related works include the use of object detection model \cite{huang2017speed} to differentiate the instances of objects \cite{lample2017unsupervised, feng2019unsupervised, laina2019towards, yao2017incorporating}. Specifically, Liu \textit{et al.} \cite{liu2019exploring} have successfully employed the weakly-supervised MIL to build the semantic concept extractor. However, these object detection schemes and MIL approaches rely heavily on expensive BBox labels.

%Instance segmentation can also be adopted to recognize object concepts, and many schemes of utilizing weaker supervision have been developed for instance segmentation \cite{khoreva2017simple}. Song \textit{et al.} \cite{song2019box} have studied the instance segmentation using BBox labels instead of the mask labels where the BBox labels are utilized to remove the irrelevant regions. Zhou \textit{et al.} \cite{zhou2018weakly} employed the supervision of image-level labels to carry out instance segmentation via back-propagating the strong visual cues. IRNet \cite{ahn2019weakly} implemented the Class Attention Maps (CAMs) as the source supervision to segment the instance masks, where the CAMs are obtained from an image classifier. 
In this research, in order to obtain rich information of objects but free the model from the bondage of expensive labels, we adopt weakly-supervised instance segmentation to recognize the object concepts, which is a primary difference from existing UIC works.

\subsection{Captioning based on Relationship Exploration}
For image captioning, many studies explore relationships between different objects to improve the performance of captioning. It is because these relationships provide rich semantic information of the input images. Among the existing works, the most popular model is graph representation which can represent the complex structural layout of both images and sentences. In supervised image captioning, Yang \textit{et al.} \cite{yang2019auto} adopted scene graph representation in the auto-encoder to obtain more human-like captions. For UIC, Gu \textit{et al.} \cite{gu2019unpaired} also applied scene graph to represent the object attributes and object relationships. Besides the scene graph representation, Cao \textit{et al.} \cite{cao2020interactions} designed a mutual attention network to reason the object-object interactions in UIC. Although various works of vision-language tasks exhibit the value of relationships between objects, it is absolutely necessary to use costly triplet annotations ($<$subject, relationship, object$>$ with locations), thereby limiting the adoption of these schemes to broader applications. 
%In visual question answering (VQA), Teney \textit{et al.} \cite{teney2017graph} proposed to employ the graph-structured representations for both scene contents and questions.

%In the proposed approach, the relationship classifier is similar in principle to \cite{baldassarre2020explanation}, which depends on BBox labels of objects and image-level relationship labels. However, we design a novel scheme using only image-level labels, properly aided by the results of weakly-supervised instance segmentation and image classification. More importantly, we improve the graph neural network (GNN) with a batch normalization (BN) scheme and the residual connection to enhance the stability of the relationship recognition model.
Distinct from these previous works, we design a novel scheme using only image-level labels to differentiate the relationships between objects, properly aided by the results of weakly-supervised instance segmentation and image classification. More importantly, a batch normalization (BN) scheme and the residual connection are adopted to enhance the stability of the basic relationship recognition model.

\section{Our Proposed Approach} 
In this section, we will first give an overview of our newly proposed weakly-supervised (WS) visual concept recognition for unpaired image captioning (WS-UIC). Then, we discuss the problem formulation of WS-UIC. After that, we will describe the WS Object Recognition (WS-OR) and WS Relationship Recognition (WS-RR) module in our framework. Finally, we present the learning of UIC model under the guidance of the aforementioned modules.

\begin{figure*}
    \centering
    \includegraphics[width=17cm, height=7.8cm]{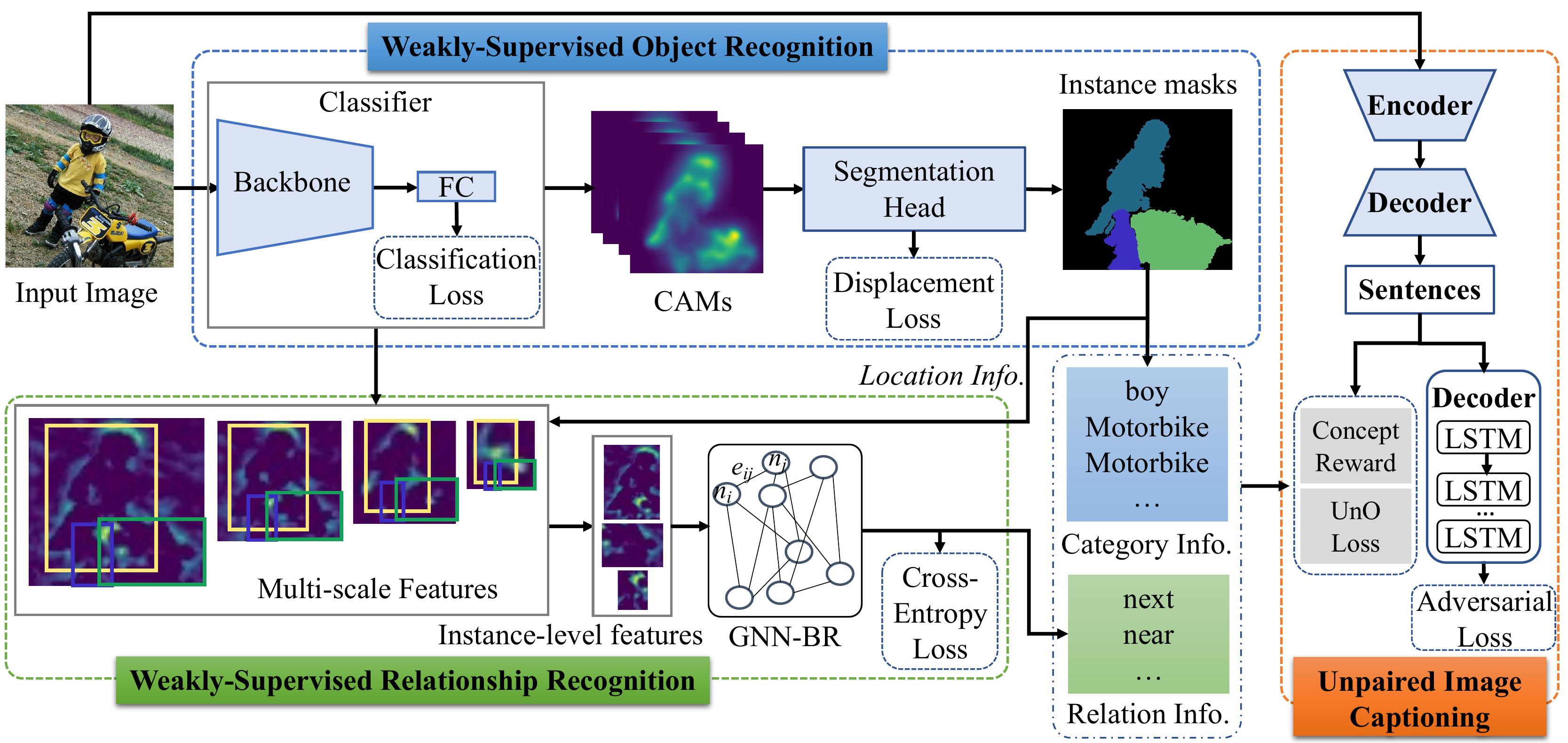}
    \caption{ \textbf{An overview of our proposed image-level weakly-supervised visual concept recognition for unpaired image captioning.} The ``Info.'' signifies ``information''. Firstly, in the weakly-supervised object recognition (WS-OR) module, an image classifier is trained using image-level labels to obtain feature maps and Class Attention Maps (CAMs). The CAMs are utilized for instance segmentation. Then the obtained instance masks are utilized together with feature maps for weakly-supervised relationship recognition (WS-RR), where an improved GNN termed GNN-BR is adopted to infer object relationships. Finally, the obtained object category and relationship information are utilized to generate image captions, where a novel loss is designed to take the unrecognized object into consideration.
    %There are two branches in the stage of visual concept recognition, \textit{i.e.}, WS-OR, and WS-RR. The ``Info.'' signifies ``information''.
    }
    \label{fig:Whole_Arc}
\end{figure*}

\subsection{Overview}
The goal of WS-UIC is to learn an image captioning model using unpaired image-text samples in a weakly-supervised manner. To achieve the goal, our newly proposed WS-UIC framework contains three main modules, \textit{i.e.}, the  WS-OR, WS-RR, and UIC model, as shown in Fig. \ref{fig:Whole_Arc}. In particular, we have a WS-OR module, consisting of an image classifier and a WS instance segmentation head, to generate the object information supervised by the image-level object labels from an image. The image classifier is trained to generate the object category information. The WS instance segmentation head, conditioned on the object category information, is utilized to obtain information of each object instance. A WS-RR module is trained by the image-level relationship labels to output the relationship concepts of an input image, where we adopt the multi-scale feature maps and spatial relations of multiple-instance to construct a graph neural network (GNN) enhanced by batch normalization and residual block. In addition, we have a UIC model to generate the image descriptions supervised by these object information and relationship concepts via a designed unrecognized object (UnO) loss integrated with concept rewards.

\subsection{Problem Formulation} 
To clearly and formally illustrate the proposed WS-UIC, we define some notations and formulate the problem of WS-UIC as a multi-stage process. Let $D_{i} = \{i^{n_i}\}_{n_i=0}^{N_i-1}$ and $D_{s} = \{s^{n_s}\}_{n_s=0}^{N_s-1}$ denote the image dataset with $N_i$ images and the sentence dataset with $N_s$ sentences, respectively. Let $D_{i,\hat{o},r} = \{(i,\hat{o},r)^{(n_{i\hat{o}r})}\}_{n_{i\hat{o}r} = 0}^{N_{i\hat{o}r} -1}$ denote the dataset with $N_{i\hat{o}r}$ tuples of images, image-level object labels, and image-level relationship labels, indexed by $i$, $\hat{o}$, and $r$, respectively. The dataset $D_{i}$ and $D_{s}$ are used for UIC, and $D_{i,\hat{o},r}$ is used for WS-OR and WS-RR, simultaneously. Formally, the goal of the WS-UIC can be written as
\begin{equation}
s \sim {\arg}\ \underset{s}{\max}\{P(s|i;\theta_{i\rightarrow s})\},
\end{equation}
where $\theta_{i\rightarrow s}$ are the model parameters to be learned in the absence of any paired $i$ and $s$, which are from independent datasets $D_i$ and $D_s$, $D_i\nleftrightarrow D_s$ ($\rightarrow$ means mapping and $\nleftrightarrow$ means independent). We use the visual concepts to learn the mapping: $i \stackrel{\theta_{i\rightarrow o}}{\longrightarrow} o\stackrel{\theta_{i,o\rightarrow r}}{\longrightarrow} r\stackrel{\theta_{i,o,r\rightarrow s}}{\longrightarrow} s$, where $o$ denotes the object instance.

According to the defined notations, the multi-stage process of the WS-UIC, \textit{i.e.}, WS-OR, WS-RR, and UIC, can be formulated as:
\begin{align}
&P(s, o, r|i) = \nonumber \\
&P(s| i,o,r; \theta_{i,o,r\rightarrow s}) & \makecell{\mathrm{UIC}} \\
&\times P(r|i,o;\theta_{i,o\rightarrow r}) & \makecell{\mathrm{WS-RR}}\\
&\times P(\hat{o} |i;\theta_{i\rightarrow o}). & \makecell{\mathrm{WS-OR}}
\label{eq:overall_framework}
\end{align}
where $P(\hat{o} |i;\theta_{i\rightarrow o})$, $P(r|i,o;\theta_{i,o\rightarrow r})$, and $P(s| i,o,r;\theta_{i,o,r\rightarrow s})$ represents the WS-OR module, WS-RR module, and UIC model, respectively.
%where $r$ represents the relationship among different objects. Given the detected object $o$, we utilize $i \leftrightarrow r$ to learn the parameters of WS-RR, \textit{i.e.}, $\theta_{i,o\rightarrow r}$.
%As shown in Fig. \ref{fig:Whole_Arc}, our proposed WS-UIC contains three main modules, \textit{i.e.}, the  WS-OR, WS-RR, and UIC model. In particular, we have a WS-OR module $P(o |i;\theta_{i\rightarrow o})$ to generate the object information $o$ supervised by the image-level object labels from an image, and a WS-RR module $P(r|i,o;\theta_{i,o\rightarrow r})$ trained using the image-level relationship labels to output the relationship concepts of the image $r$. In addition, we have a UIC model $P(s| i,o,r;\theta_{i,o,r\rightarrow s})$ to generate the image descriptions supervised by these object information and relationship concepts.
In the \textit{inference} phase, only the UIC model is needed to describe an unseen image:
\begin{equation}
s \sim {\arg}\ \underset{s}{\max}\{P(s| i; \theta_{i,o,r\rightarrow s})\}.
\end{equation}

In the following subsections, we will depict the WS-OR, WS-RR, and UIC modules in a more detailed way. It is worth noting that only image-level labels are adopted in WS-UIC to train the WS-OR and WS-RR module in a weakly-supervised manner, but other works depend on enormous expensive labels, such as relationship-triplet labels, to obtain the object or/and relationship information in an image.

\subsection{Exploring Weakly-Supervised Object Recognition}\label{sec:object}
For UIC, image-text pairs are not available, therefore, the object concepts of images are taken as the crucial clues for generating accurate captions. Previous works utilize object detection models to identify these object concepts but rely on enormous Bounding Box (BBox) labels (as shown in Fig. \ref{fig:Rel_labels} (b)) to train the model. To overcome this drawback, a WS-OR model $P(\hat{o} |i;\theta_{i\rightarrow o})$ is proposed to recognize the object concepts given only image-level labels, as shown in Fig. \ref{fig:Rel_labels} (d). The main structure of WS-OR includes an image classifier and an image-level instance segmentation module \cite{ahn2019weakly}. Therefore, utilizing only \textit{image-level} labels is capable of differentiating individual objects.

\noindent \textbf{Image classification} plays essential roles in our proposed approach: 
First, it can enlarge the number of object categories of the captioning dataset to increase the vocabulary of the UIC model. 
Second, it can be implemented to generate the feature maps $\textbf{\textit{F}}$ and Class Attention Maps (CAMs) \cite{zagoruyko2016paying} which define the distinct areas of object instances and thus can be adopted to train the image-level instance segmentation model. 
Without loss of generality, in our experiments, we select the ResNet50 \cite{he2016deep} as the backbone of the classification network. And we adopt the popular used multi-label soft margin loss \cite{yang2018sgm}. To get the CAMs \cite{zagoruyko2016paying}, we follow \cite{zhou2016learning} by using the Global Average Pooling (GAP) in Convolutional Neural Networks (CNNs). Formally, the CAM of an object class $\hat{o}_{i}$ is denoted by $M_{\hat{o}_{i}}$,
\begin{equation}
M_{\hat{o}_{i}}(x,y) = \frac{\phi_{\hat{o}_{i}}^{T}f(x,y)}{\max_{x,y}\phi_{\hat{o}_{i}}^{T}f{(x,y)}},
\label{eq:cam}
\end{equation}
where $f{(x,y)}$ represents the feature map with coordinate $(x,y)$, and $\phi_{\hat{o}_i}$ is the classification weights of object class $\hat{o}_{i}$.

\begin{figure}[!tb]
     \centering
     \includegraphics[width=8cm, height=7.5cm]{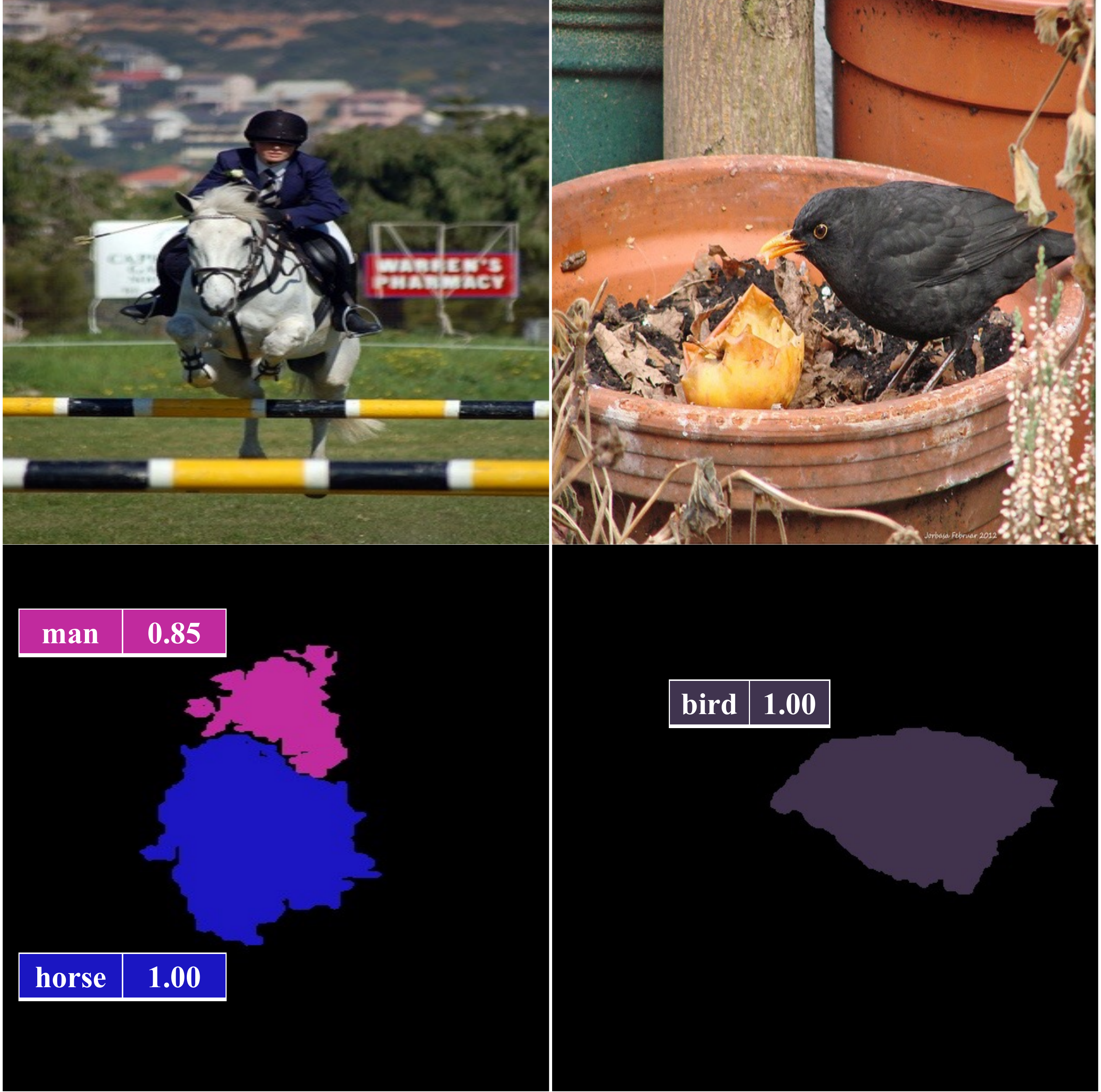}
     \caption{The input images and their WS-OR results, including the instance masks $\textbf{m}_i$, object class $o_i$, and confident score $v_{i}$ .}
     \label{fig:Seg_results}
 \end{figure}

\noindent \textbf{Image-level instance segmentation} is another critical component of WS-OR. It is utilized to recognize instance information for individual objects in an image, including instance category $o_i$, instance mask $\textbf{m}_i$, and the corresponding confident score $z_i$. We follow IRNet \cite{ahn2019weakly} to address the instance segmentation issue by adopting the CAMs as the source supervision. Fig. \ref{fig:Seg_results} illustrates some results of the WS-OR module.

The obtained object information can be formulated as below: 
\begin{equation} 
\mathcal{O} = [{(o_1, \textbf{m}_1, z_{1}),...,(o_i ,\textbf{m}_i, z_{i} ),...,(o_{N_{\mathcal{O}}} ,\textbf{m}_{N_{\mathcal{O}}}, z_{N_{\mathcal{O}}} )}],
\end{equation}
where $N_{\mathcal{O}}$ is the total number of object concepts in an image. The obtained object information is utilized to predicate the relationships between objects in Section~\ref{sec:relation} and train the UIC model to generate descriptions in Section~\ref{sec:UIC}. It is worth to mention that we can adopt any  \textit{image-level} instance segmentation approach or \textit{image-level}  object detection scheme to differentiate objects. 
%as long as only image-level labels are utilized to supervise it.

\subsection{Exploring Weakly-Supervised Relationship Recognition}\label{sec:relation}
\begin{figure}[!tb]
\small
\centering
\includegraphics[width=8cm, height=8cm]{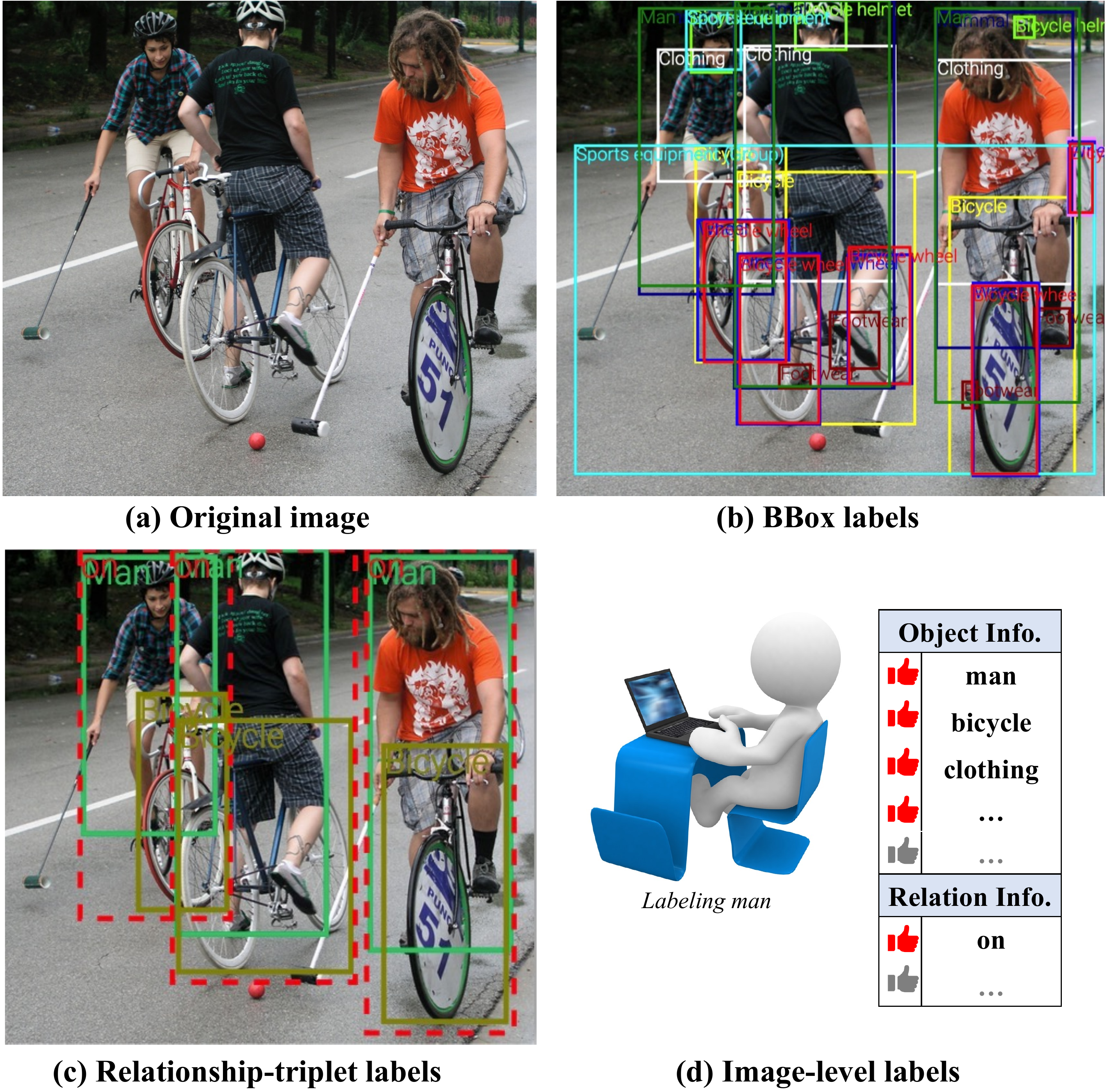}
\caption{ \textbf{An example of BBox labels, relationship-triplet labels, and image-level labels.} Both image-level object labels and image-level relationship labels belong to image-level labels. (a), (b), (c), and (d) are the original image, the BBox labels, the relationship-triplet labels, and the image-level labels, respectively. The BBox labels present the category of each object instance with a location. The relationship labels provide the category of each (subject, relationship, object) instance with three locations. And the image-level labels only indicate the concept categories without repetition for each instance and the location information. The annotation cost of BBox labels and relationship-triplet labels is much higher than image-level labels.}
\label{fig:Rel_labels}
\end{figure}

In addition to the object category information, the relationships between these objects are another crucial part of the image captions. Most existing schemes \cite{baldassarre2020explanation, gu2019unpaired} explore these relationship concepts by using enormous relationship-triplet labels, which are \textit{expensive to be annotated and hard to collect} in practice. To address this issue, we propose utilizing image-level relationship labels to train the model to recognize the relationship concepts. Fig. \ref{fig:Rel_labels} (c) and (d) show an example of relationship-triplet labels and image-level relationship labels respectively. 

\begin{figure}[!tb]
    \centering
    \includegraphics[width=8cm, height=6cm]{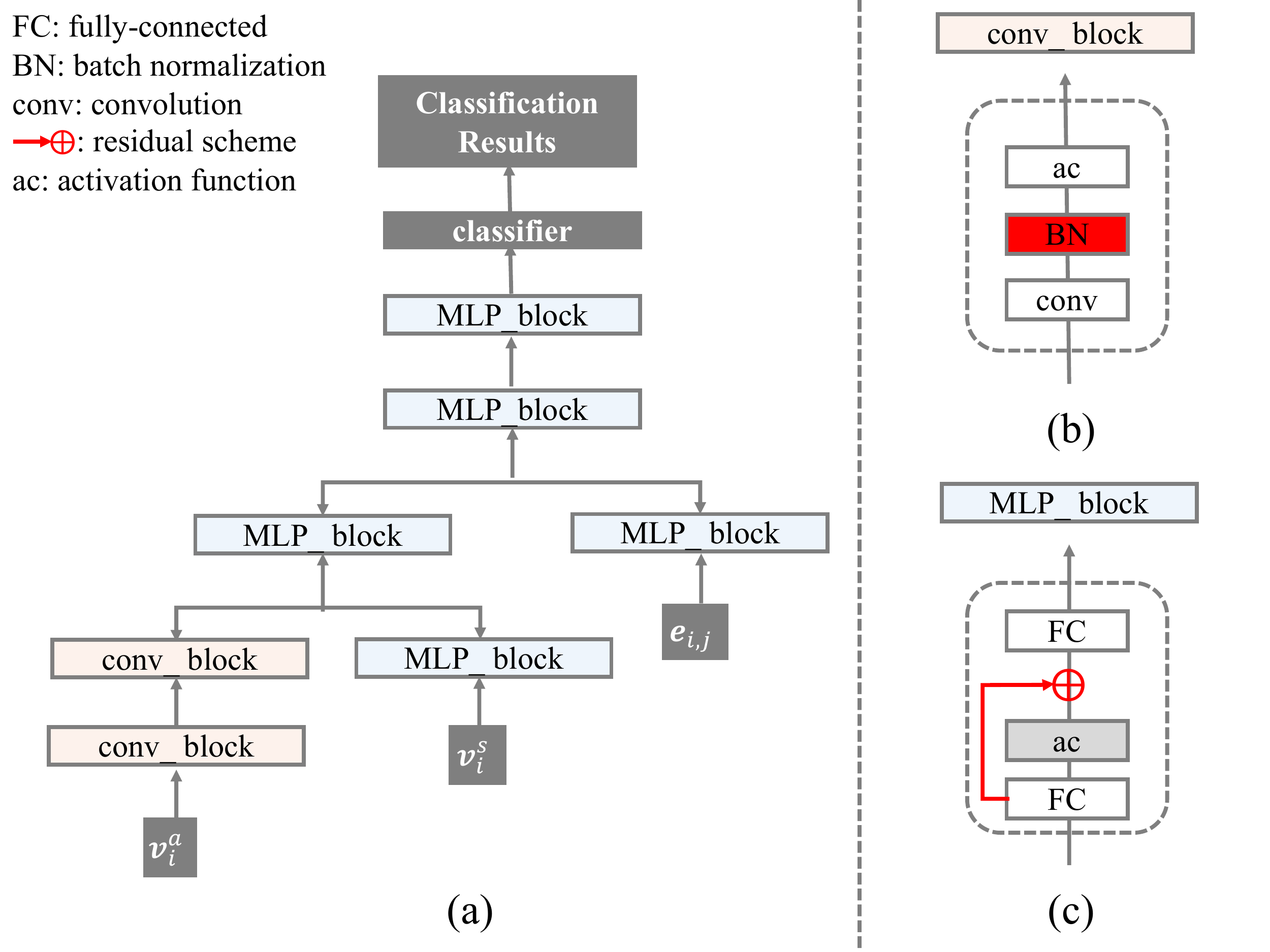}
    \caption{ \textbf{The detailed network architecture of GNN-BR.} (a) is the overall architecture of GNN. (b) and (c) are the enhanced ``conv layer" with BN scheme and ``MLP layer" with residual scheme separately.}
    \label{fig:mlp_gnn}
\end{figure}

In our framework, the GNN is adopted to model the relations between multiple instances in an image. The GNN takes the graph representation of an image $\mathcal{G} = (\mathcal{V}, \mathcal{E})$ as input and predict the relationships among objects. As the image-level relationship labels contain no object information, therefore, we need to use the object information $\mathcal{O}$ obtained from Section~\ref{sec:object} to construct the graph representation. Specifically, the nodes $\textit{\textbf{v}}_i = (\textit{\textbf{v}}_i^a, \textit{\textbf{v}}_i^s)$, indicating an object with its appearance features and spatial features, and edges $\textbf{e}_{ij}$ of the graph are transformed through small networks separately. Then pairs of nodes and the edge connecting them are input into a several Multi-Layer Perceptron (MLP) network \cite{baldassarre2020explanation}. The graph is fully-connected since the relationships between all pairs of objects are considered. Later, a classier is applied to each of the edges, and the softmax outputs are supervised by the image-level relationship labels via a cross-entropy loss. The whole architecture is illustrated in Fig. \ref{fig:mlp_gnn} (a).

To make the training of our model more stable and easier to converge, the BN and residual schemes are introduced in GNN, as shown in Fig. \ref{fig:mlp_gnn} (b) and (c) respectively. The BN with fixed means and expectations is adopted in the convolutional layer, which keeps the features from different layers following the same suitable distribution. It makes the network less sensitive to the initialization and can accelerate the convergence speed \cite{ioffe2015batch}. Formally,
\begin{equation}
\hat{\textbf{h}}^{k} = \frac{\textbf{h}^{k}-E[\textbf{h}^{k}]}{\sqrt{Var[\textbf{h}^{k}]}},
\end{equation}
where $\textbf{h}$ indicates the output of the former layer, $k$ is the dimensional number of $\textbf{h}$. $E$ and $Var$ mean the expectation and variance, respectively. 
The residual block is designed to solve the gradient degradation problem via adding the output of former layers with current layers \cite{he2016deep}. Formally,
\begin{equation}
\hat{F}(\textbf{h}) = F(\textbf{h}) + \textbf{h},
\end{equation}
where $F(\textbf{h})$ is the output of the current layer. We adopt the residual scheme in each MLP block of the WS-RR network.
 
Through this section, we can obtain the relationship concepts of images:
\begin{equation}
\mathcal{R} = [{\left(r_1, z_{1}\right),...,\left(r_k, z_{k} \right),...,\left(r_{N_{\mathcal{R}}}, z_{N_{\mathcal{R}}} \right)}], 
\end{equation}
where $r_k$ is the $k$-th inferred relationship class, $z_k$ represents the corresponding confidence score, and $N_{\mathcal{R}}$ defines the total number of the recognized relationship concepts. The recognized relationship information can be utilized to guide UIC on generating relevant image captions in Section~\ref{sec:UIC}.
% \begin{figure}[!htb]
% \centering
%\subfloat[]{
% \includegraphics[width=2cm,height=1.93cm]{./Images/chap3/Pred_input.png}
% }
% \quad
%\subfloat[]{
% \includegraphics[width=2cm,height=1.93cm]{./Images/chap3/fms.png}
% }
% \quad
%\subfloat[]{
% \includegraphics[width=2cm,height=1.93cm]{./Images/chap3/Pred_seg.png}
% %\caption{fig1}
% }
% \quad
%\subfloat[]{
% \includegraphics[width=2cm,height=1.93cm]{./Images/chap3/Pred_Bbox.png}
% }
% \quad
%\subfloat[]{
% \includegraphics[width=3cm,height=1.24cm]{./Images/chap3/multi_scale_fms.png}
% }
% \quad
%\subfloat[]{
% \includegraphics[width=0.4cm,height=1.5cm]{./Images/chap3/object_level_fm.png}
% }
% \caption{The generation process of object-level appearance features. The set of images are (a) an input image, (b) the feature maps of the input image, (c) the instance segmentation masks, (d) the instance Bboxes, (e) multi-scale feature maps with Bboxes, and (f) the object-level appearance features of different instances.}
% \label{fig:obj_level_features}
% \end{figure}

\subsection{Unpaired Image Captioning}\label{sec:UIC}
Similar to the standard image captioning models, the UIC model adopts the encoder-decoder framework \cite{wu2019recall} to encode an image into features and decode these features into captions. Any CNN backbone networks can be used for the feature extraction. In our experiments, we select the inception-V4 \cite{szegedy2017inception} as the image encoder, and obtain the feature representation $\textbf{f}_{im}$. For the decoder network, we adopt the popular Long-short Term Memory (LSTM) network \cite{zhang2018high} which takes the encoded feature maps as input at the first time step. Then, it recurrently takes its hidden state and the word embedding of the previous time step as the input and output one word and the hidden state for subsequent time steps \cite{xu2019multi}.

Unlike traditional captioning models, the ground-truth captions are not available for UIC, therefore, we have to find other useful information for the optimization of our network. Firstly, a discriminator with the LSTM architecture is designed to differentiate the true sentences and generated sentences via an adversarial loss \cite{feng2019unsupervised}. Secondly, since the objects and the relationships of objects are the key components of a caption, they can be served as the supervision for UIC. In this work, we leverage both the concept reward loss and unrecognized object loss to train the model. More details of the two losses are given below.

\noindent \textbf{Concept Reward.} The recognized object concepts and relationship concepts $\mathcal{W} \in ({\mathcal{O}, \mathcal{R}})$ provide image-words pairs for UIC. If the $t$-th generated word $s_t$ is in the recognized visual concepts, a concept reward $R_t$ will be assigned to the $t$-th generated word $s_t$ as follows:
\begin{equation}
R_t = \underbrace{\alpha \times \sum_{i=1}^{N} \textbf{I}\left(s_t = o_i\right) \times z_{i}}_{\text{object concept reward}} + \underbrace{\beta \times \sum_{k=1}^{K} \textbf{I}\left(s_t = r_k\right) \times z_{k}}_{\text{relationship concept reward}} ,
\label{equ:concept_reward}
\end{equation}
where the $\alpha$ and $\beta$ are the weights of object concept rewards and relationship concept rewards, respectively. $\textbf{I}(\cdot)$ is the indicator function \cite{kenny2003indicator}.
%The generated captions might contain the object concepts that does not recognized by the WS-OR ($q_i \notin \mathcal{O}$) but the $q_i$ (\textit{e.g.} 'umbrella') might related to the recognized object concepts (\textit{e.g.} 'rain') or not related to the recognized object concepts (\textit{e.g.} 'sunny'). Thus $q_i$ has a positive or negative influence on the captions, which motivates us to design an unrecognized object loss to supervise the UIC model. The loss value $z_i$ is learned by the network. If the $q_i$ has a positive effect on the caption, its score $z_i$ will be less than 0. Otherwise, $z_i > 0$. Formally,
\begin{figure}[!tb]
\small
\centering
\includegraphics[width=8.8cm, height=9cm]{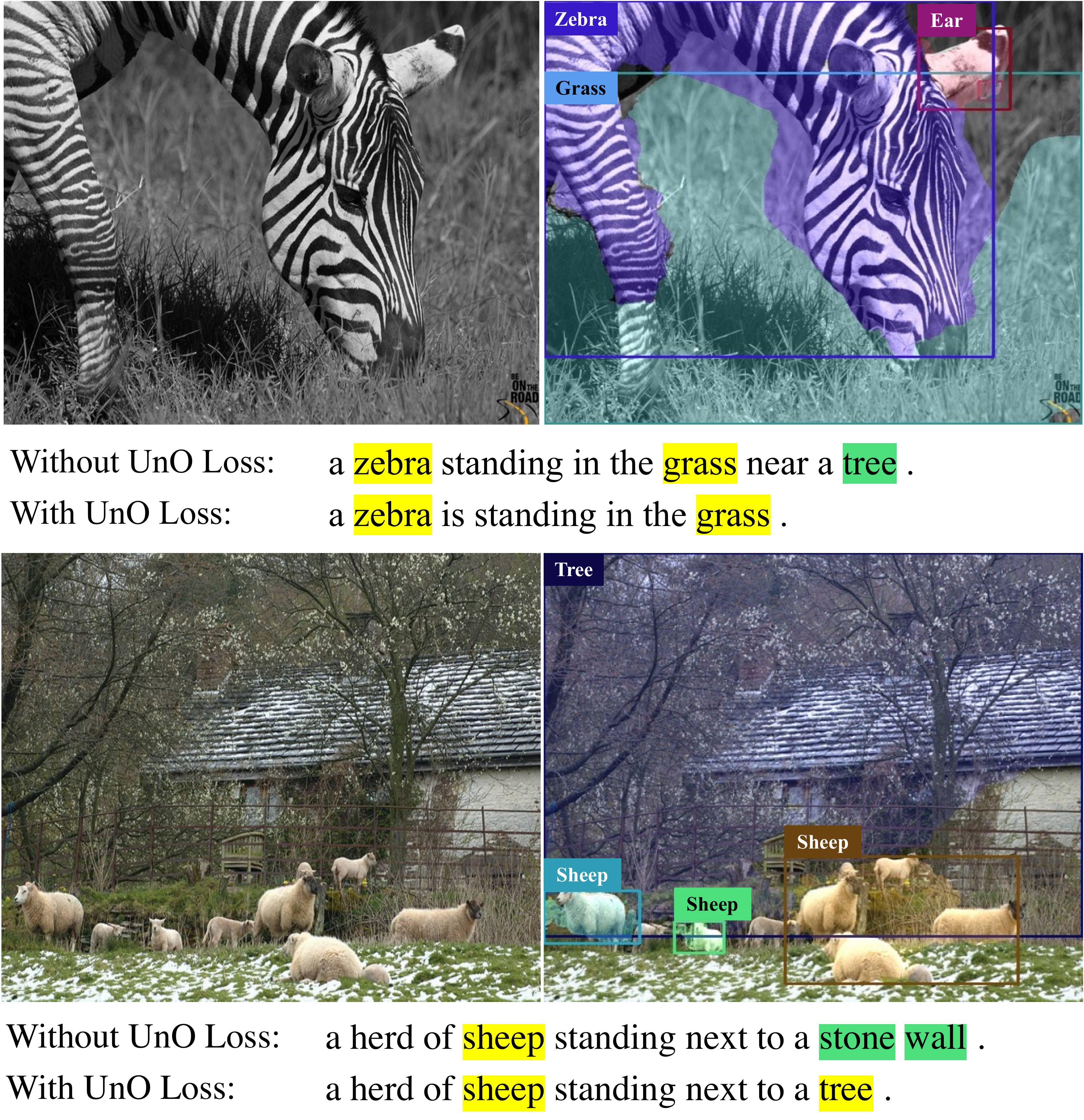}
\caption{\textbf{Two examples of the unrecognized objects presenting in the captions.} The left and right images are the input images and the recognized objects, respectively. Below the images, two types of captions are listed respectively, \textit{i.e.}, without UnO and with UnO loss. } 
\label{fig:UoL}
\end{figure}

\noindent \textbf{Unrecognized Object Loss.} The generated captions might contain the object concepts that are not recognized by the WS-OR ($q_i \notin \mathcal{O}$) model. As illustrated in Fig. \ref{fig:UoL}, the object categories \emph{tree}, \emph{stone}, and \emph{wall} are not recognized by the WS-OR model, but presenting in the captions generated by the UIC model. These unrecognized objects motivate us to design an unrecognized object loss for the UIC model. The loss value $z_i$ of $q_i$ is learned by the network. Formally,
\begin{equation}
L_t^{u} = \underbrace{\lambda \times \sum_{i=1(q_i \notin \mathcal{O})}^{N_q} \textbf{I}\left(s_t = q_i\right) \times z_{i}}_{\text{unrecognized object loss}},
\label{equ:concept_penalty}
\end{equation}
where $\lambda$ is the weight of the loss and $N_q$ is the total number of unrecognized objects.
The loss for $s_t$ is a combination of the visual concept reward and unrecognized object loss: 
\begin{equation}
L_t = - R_t + L_t^{u}.
\label{equ:concept_punishment}
\end{equation}

% Besides the loss related to visual concepts, the image feature reconstruction loss and sentence reconstruction loss \cite{feng2019unsupervised} are also utilized to train the UIC model. In addition, we adopt a convolutional block attention module \cite{woo2018cbam} to refine the image features.

\subsection{Training and Inference Phase} 

In this paper, we adopt multi-stage optimizations for our network since the proposed framework contains multiple components that are needed to be optimized independently. 

\noindent \textbf{Training Phase.} Our network can be optimized via the following three stages:   

\textbf{\emph{Stage-I: }} The WS-OR model is first trained by a dataset with a large number of object categories with the image-level object labels. The image classification model is optimized by a multi-label soft margin loss \cite{yang2018sgm}. The images in the training dataset of captioning are inferenced by the image classification model to obtain the object categories, feature maps, and CAMs. Later, the object information is adopted to train the image-level instance segmentation with the optimization of a displacement loss \cite{ahn2019weakly}, from which the instance masks of objects can be obtained.

\textbf{\emph{Stage-II: }} Aided by the aforementioned object categories and object instance masks, WS-RR model is trained to recognize the relationships between objects, which is optimized by a cross-entropy loss of image-level relationship labels.

\textbf{\emph{Stage-III: }} The object concepts and relationship concepts are utilized to train the UIC model which is optimized by the object reward and the unrecognized object loss. Finally, the generated captions are utilized as pseudo captions to train the fully-supervised image captioning model with a signal cross-entropy loss. These captions, which contain at least one recognized object concept, are filtered out as pseudo captions.

\noindent \textbf{Inference Phase.} 
It is worthy to note that only the image captioning model obtained from the multi-stage training phase is used in the inference phase. That is to say, we do not have to recognize the objects and the relationship concepts. Given a testing image, we can directly generate its captions using the captioning model.

\section{Experiments}
% In this section, we evaluate the effectiveness of the proposed Weakly-Supervised (WS) visual concept recognition for unpaired image captioning (WS-UIC). The datasets and the implementation settings are presented firstly. Later, we elaborate the experimental results by comparing them with the state-of-the-art approaches that require stronger supervision under unpaired settings. Then, we compare the labeling cost of related approaches. Intensive ablation studies are also reported to demonstrate the effectiveness of each component of our WS-UIC, and some qualitative results are presented. Finally, we show some bad cases and discuss the possible reasons and future works. 
In this section, we evaluate the effectiveness of the proposed Weakly-Supervised (WS) visual concept recognition for unpaired image captioning (WS-UIC).

\subsection{Dataset and Evaluation Metric} 

\noindent \textbf{Dataset. }
For WS visual concept recognition, we use the Visual Genome (VG) dataset~\cite{krishna2017visual} to train the WS object recognition (WS-OR) and WS relationship recognition (WS-RR) models, where 305 object categories and 64 relationship categories are utilized. Then the COCO images are adopted to derive the object and relationship concepts. Only the image-level class labels are utilized in these procedures. For UIC, we utilize COCO images paired with COCO captions to train the model \textit{in an unpaired way}. We adopt the common splits of COCO images \cite{karpathy2015deep}: 113,287 training images, 5000 validation images, and 5000 test images. The COCO captions are also utilized to construct the vocabulary corpus with words presenting no less than 4 times. 

To illustrate the robustness of visual concept recognition of WS-UIC, we use another dataset to train the WS-OR model and the WS-RR model, \textit{i.e.}, OpenImage V2 \cite{krasin2017openimages} with 545 object categories. Shutterstock sentence corpus \cite{feng2019unsupervised} is implemented as another sentence corpus, and COCO images in the training set are taken as the image corpus in the experiments. The sentence corpus contains 2,282,444 distinct image descriptions.

\noindent \textbf{Evaluation Metric.}
We report the BLEU-1 (B1), BLEU-2 (B2), BLEU-3 (B3), BLEU-4 (B4) \cite{papineni2002bleu}, Meteor (M) \cite{denkowski2014meteor}, ROUGE (R) \cite{lin2004looking}, CIDEr (C) \cite{vedantam2015cider} and SPICE (S) \cite{anderson2016spice} scores which are calculated via the ground-truth captions of the test images. Among these metrics, BLEU and Meteor (M) were produced for machine translation in 2002 and 2014, ROUGE (R) was mainly proposed for automatic abstracting in 2004, CIDEr (C) and SPICE (S) were mainly for image captioning proposed in 2015 and 2016, respectively.

\subsection{Implementation Details}
For WS-OR, we select the object categories for an image with logits larger than 2 in the image classification stage. The ResNet50 \cite{he2016deep} is taken as the backbone. Then, we follow IRNet \cite{ahn2019weakly} to tackle image-level instance segmentation.

In the WS-RR stage, the batch size is set as 256, the relationships with confident scores larger than 0.7 will be filtered out to train the UIC model. We choose GeLU \cite{tolstikhin2021mlp} as the activation function. Four scales of image-level feature maps are used in our experiments.

For the UIC model, we set both the LSTM hidden dimension and the shared latent space dimension as 512. The learning rates are set to be 0.00001. We also adopt the initialization pipeline following \cite{feng2019unsupervised}. The weighting hyper-parameters of the concept reward and unrecognized object loss are chosen roughly at the same scales. Specially, $\alpha$, $\beta$ and $\lambda$ are set as 1, 0.5, 1, separately. The code of this paper is developed based on TensorFlow framework, and all the experiments are conducted on a server with 4 V100 GPUs.

\subsection{Comparison on Benchmarks}

Three types of experimental comparisons are conducted in the following parts. The first one is to compare the WS-UIC with Pivoting \cite{gu2018unpaired}, Song \textit{et al.} \cite{song2019unpaired}, Guo \textit{et al.} \cite{guo2019mscap}, Feng \textit{et al.} \cite{feng2019unsupervised}, Laina \textit{et al.} \cite{laina2019towards}, Cao \textit{et al.} \cite{cao2020interactions}, Gu \textit{et al.} \cite{gu2019unpaired}, SCS \cite{ben2021unpaired}, illustrated in Table \ref{tab:perf_com_coco_many_VG}. These approaches implement the same dataset at the UIC stage. The second one is to compare the WS-UIC with Feng \textit{et al.} \cite{feng2019unsupervised} with the same dataset settings at both the visual concept recognition stage and the UIC stage, shown in Table \ref{tab:perf_com_coco_Feng}. The third one is to compare the WS-UIC with Feng \textit{et al.} \cite{feng2019unsupervised} via implementing the independent datasets, \textit{i.e.}, COCO images with Shutterstock sentences.
%\begin{table}[!t]\small
%    \centering
%    \setlength{\tabcolsep}{1mm}{
%    \begin{tabular}{ccccccccc}
%    \hline
%     \multirow{1}{*}{Supervision} & \multirow{1}{*}{Method}  &Label type  &  Number\\ \hline
%        
%    \multirow{8}{*}{\textit{Fully}} &  \makecell{ Feng \textit{et al.}  \cite{feng2019unsupervised} }                   && \\
%     & \makecell{ Laina \textit{et al.}  \cite{laina2019towards}  }   &&\\  
%& \makecell{Pivoting \cite{gu2018unpaired}}       &&\\
%      & \makecell{Song et al. \cite{song2019unpaired}}              &&\\ 
%     & \makecell{Guo et al. \cite{guo2019mscap}}            &&\\ 
%     & \makecell{Cao \textit{et al.} \cite{cao2020interactions}}   & &\\   
%     & \makecell{Graph-Align \cite{gu2019unpaired}}  &&\\  
%     & \makecell{SCS \cite{ben2021unpaired}}   & &\\ \hline 
%     % \multicolumn{6}{c}{\textit{Weakly-Supervised Setting}}\\
%%\textit{Weakly}   &   WS-UIC           & 20.5&20.0&45.5&64.1&13.4\\  \hline
%\textit{Weakly} & \makecell[c]{WS-UIC}&&\\  \hline
%    \end{tabular}}
%    \caption{Label difference in different approaches.}
%        \label{tab:label_difference}
%\end{table}

\begin{table}[!t]\small
    \centering
    \caption{Performance comparisons on the test split of the COCO dataset under the unpaired setting. All these referenced approaches are fully-supervised in the visual concept recognition stage utilizing millions of expensive labels, such as relationship-triplet labels, image-caption labels, and object attribute labels, as illustrate in Table \ref{tab:label_difference}.}
    \setlength{\tabcolsep}{1mm}{
    \begin{tabular}{c|ccccccc}
    \hline
     \multirow{1}{*}{Supervision} & \multirow{1}{*}{Method}  &B4 & M & R & C  &  S\\ \hline
        %\multicolumn{6}{c}{\textit{Fully-Supervised Setting}} \\
    \multirow{9}{*}{\textit{Fully-supervised}} &  \makecell{Pivoting \cite{gu2018unpaired}}        &5.4&13.2&-&17.7&-\\
      & \makecell{Song \textit{et al.} \cite{song2019unpaired}}             &11.1&14.2&-&28.2&-\\ 
      & \makecell{Con2sen \cite{feng2019unsupervised}} &11.3&15.7&37.9&33.9&9.1\\ 
     & \makecell{Guo \textit{et al.} \cite{guo2019mscap}}         &-&16.8&-&55.3&-\\ 
   & \makecell{ Feng \textit{et al.}  \cite{feng2019unsupervised} }       &18.6&17.9&43.1&54.9&11.1\\   
     & \makecell{ Laina \textit{et al.}  \cite{laina2019towards}  } & 19.3&20.2&45.0&61.8&12.9\\  \cline{2-7}
     & \makecell{Cao \textit{et al.} \cite{cao2020interactions}}  & 21.9&21.1&46.5&64.0&14.5\\   
     & \makecell{Gu \textit{et al.} \cite{gu2019unpaired}}   &21.5&20.9&47.2&69.5& 15.0\\  
     & \makecell{SCS \cite{ben2021unpaired}}   &  22.8&21.4&47.7&74.7&15.1\\ \hline
     % \multicolumn{6}{c}{\textit{Weakly-Supervised Setting}}\\
%\textit{Weakly}   &   WS-UIC           & 20.5&20.0&45.5&64.1&13.4\\  \hline
\textit{Weakly-supervised} & \makecell[c]{WS-UIC}& 21.5  &20.1&45.8&65.7& 13.6\\  \hline
    \end{tabular}}
        \label{tab:perf_com_coco_many_VG}
\end{table}

As shown in Table \ref{tab:perf_com_coco_many_VG}, we report our results and compare with other state-of-the-art methods, including \cite{ gu2018unpaired, song2019unpaired, guo2019mscap, feng2019unsupervised, laina2019towards, cao2020interactions, gu2019unpaired, ben2021unpaired}. Note that all these methods use the same dataset, \textit{i.e.}, the COCO dataset, for the training of the captioning model. From the Table, we can find that the performance of Cao \textit{et al.} \cite{cao2020interactions}, Gu \textit{et al.} \cite{gu2019unpaired}, and SCS \cite{ben2021unpaired} are slightly better than ours. The reason is that these methods adopt fully-supervised methods for visual concept recognition, that is to say, many BBox and relationship-triplet labels are used for training which are very expensive to annotate. Therefore, our proposed image-level weakly-supervised methods for UIC is a more cost-effective approach. On the other hand, our proposed method is significantly better than \cite{gu2018unpaired, song2019unpaired, feng2019unsupervised, guo2019mscap, laina2019towards}, although these methods are also developed based on millions of much more costly labels, as shown in Table \ref{tab:label_difference}.

\begin{table}[!t]\small
    \centering
    \caption{Performance comparisons of WS-UIC and the state-of-the-art approaches that leveraged the OpenImage V2 as the dataset of visual concept recognition.}
    \setlength{\tabcolsep}{1mm}{
    \begin{tabular}{ccccccc}
    \hline
   \multirow{1}{*}{\makecell{Supervision}} & \multirow{1}{*}{Method} & B4 & M & R & C  &  S\\ \hline
     \multirow{2}{*}{\makecell{\textit{Fully-supervised}} }& \makecell{Con2sen \cite{feng2019unsupervised}} &11.3&15.7&37.9&33.9&9.1\\ 
    & \makecell{Feng \textit{et al.} \cite{feng2019unsupervised}} &18.6&17.9&43.1&54.9&11.1\\ \hline
 \multirow{1}{*}{\makecell{ \textit{Weakly-supervised}}}  &  \makecell[c]{WS-UIC w/o \\ relationship}  &19.0&18.6&43.8&55.7&11.4\\ \hline  
% &  \makecell[c]{WS-UIC}  && &&&\\ \hline
  %     WS-UIC& \XSolidBrush  &&&&&\\   \hline
    \end{tabular}}
    
    \label{tab:perf_com_coco_Feng}
\end{table}

\begin{table}[!t]\small
    \centering
    \caption{Performance comparisons on the WS-UIC with state-of-the-art approaches for unpaired captioning with independent data (COCO images $\leftrightarrow$ Shutter stock sentence). ``Sup." means supervision.}
    \setlength{\tabcolsep}{1mm}{
    \begin{tabular}{cccccccccc}
    \hline
   \multirow{1}{*}{\makecell{Sup.}} & \multirow{1}{*}{Method} & B1 & B2& B3 &B4 & M & R & C  &  S\\ \hline
     \multirow{2}{*}{\makecell{\textit{Fully}\\\textit{-Sup.}} }& \makecell{Con2sen \cite{feng2019unsupervised}} &37.2&20.0&9.6&4.7&12.3&27.3&22.5&8.2\\ 
    & \makecell{Feng \\ \textit{et al.} \cite{feng2019unsupervised}} &41.0& 22.5 & 11.2 &5.6&12.4&28.7&28.6&8.1\\ \hline
 \makecell[c]{\textit{Weakly}\\\textit{-Sup.}}  &  \makecell[c]{WS-UIC}  & 41.3 & 22.4 &11.4&5.9&12.0&28.0&26.9&7.6\\ \hline 
% &  \makecell[c]{WS-UIC}  && &&&\\ \hline
  %     WS-UIC& \XSolidBrush  &&&&&\\   \hline
    \end{tabular}}
    \label{tab:perf_com_coco_website}
\end{table}

In addition, we compare our model with Feng \textit{et al.} \cite{feng2019unsupervised} by leveraging the same dataset settings at all stages. Specifically, we all adopt the OpenImage V2 with the same 545 object categories for the visual concept recognition, and utilize the COCO dataset for the training of UIC model. For the sake of fairness, we removed the WS-RR module and compared it with \cite{feng2019unsupervised} under the same configuration. As shown in Table \ref{tab:perf_com_coco_Feng}, we can achieve better performance than Feng et al. \cite{feng2019unsupervised} under all five evaluation metrics. More importantly, our model requires zero BBox labels, however, almost 0.7 million BBox labels are leveraged for the training of \cite{feng2019unsupervised}. We also compare with Feng et al. \cite{feng2019unsupervised} by adopting the independent datasets, \textit{i.e.}, COCO images with Shutterstock sentences. As illustrated in Table \ref{tab:perf_com_coco_website}, our method is also better than theirs considering the almost two times of labeling cost of \cite{feng2019unsupervised}, even under the more challenging unpaired settings. All these experimental results and analyses demonstrate the effectiveness and advantages of our model.

\begin{table}[!t]\small
    \centering
    \caption{Ablation study to evaluate the effectiveness of each component in WS-UIC. The ``Obj" means using the object concepts to train the UIC model. The ``Rel" indicates using the relationship concepts to train the UIC model. The ``UnO" represents leveraging the unrecognized object loss to train the UIC model. The WS-UIC means adopting all these schemes together to train the UIC model and is optimized by the pseudo captions finally.} %The image and sentence datasets for image captioning are unpaired.
    \setlength{\tabcolsep}{1mm}{
    \begin{tabular}{c|c|c|c|c|cc|c|c} \hline
    \multirow{2}{*}{Approach}  & \multicolumn{5}{c}{Evaluation Metrics} \\ \cline{2-6}
         & B4 & M & R & C  &  S\\ \hline
     Obj   &19.3&19.3&43.8&60.7&12.9 \\  \hline
     %Obj+Att   &19.5&19.4&44.4&61.7& 12.9\\  \hline
     \makecell[c]{Obj + Rel} &19.7 &19.6 &44.5 &62.1&13.0 \\  \hline
     Obj + UnO  &19.9&19.6&44.8&61.8&13.0 \\  \hline
     Obj + UnO + Rel &20.5&20.0&45.5&64.1&13.4 \\  \hline
     %\makecell[c]{WS-UIC}&20.5 &20.0&45.5&64.1& 13.4\\  \hline
     
     \makecell[c]{WS-UIC}&\textbf{21.5}  &\textbf{20.1}&\textbf{45.8}&\textbf{65.7}& \textbf{13.6}\\  \hline
    \end{tabular}}
    \label{tab:ablation_study}
\end{table}

\subsection{Comparisons on the Labeling Cost of Related Approaches}
\begin{table}[!t]\small
    \centering
    \caption{The number of mainly additive labels, \textit{i.e.}, image caption pairs, BBox labels,  relationship-triplet labels, and image-level labels utilized in different approaches. All of the previous approaches depended on millions of at least one of the former three expensive labels. But, WS-UIC requires only cheap image-level labels.} %The image and sentence datasets for image captioning are unpaired.
    \setlength{\tabcolsep}{1mm}{
    \begin{tabular}{c|c|c|c|c|c} \hline
    \multirow{2}{*}{Approach}  & \multicolumn{4}{c|}{\# Additive Labels (M)} & \multirow{2}{*}{\makecell{Cost\\ (Dollar)}} \\ \cline{2-5}
         & \makecell{Image-\\caption pair}& BBox & \makecell{Relationship\\-triplet} & \makecell{Image\\-level} \\ \hline
     Pivoting \cite{gu2018unpaired}   &0.24&0&0 & 0 & -\\  \hline
     %Obj+Att   &19.5&19.4&44.4&61.7& 12.9\\  \hline
     \makecell[c]{Song \textit{et al.} \cite{song2019unpaired}} &1.82 &0&0 & 0 & -\\  \hline
     Guo \textit{et al.} \cite{guo2019mscap}  &0.12 &0&0 & 0 & -\\  \hline
     Feng \textit{et al.} \cite{feng2019unsupervised} &0&0.71 &0 & 0 & 76.7k\\  \hline
     Laina \textit{et al.} \cite{laina2019towards}&0&1.05 &0 & 0 &113.4k \\  \hline
     Cao \textit{et al.} \cite{cao2020interactions}&0&0.28 &0.20 & 0 & 95.0k \\  \hline
     Gu \textit{et al.} \cite{gu2019unpaired}&0&3.84 &2.35 & 0 & 1176.1k\\  \hline
     SCS \cite{ben2021unpaired}&0&3.84&0 & 0 & 414.7k \\\hline
     %\makecell[c]{WS-UIC}&20.5 &20.0&45.5&64.1& 13.4\\  \hline
     
     \makecell[c]{WS-UIC}&0  &0&0& 1.11 & 40.0k\\  \hline
    \end{tabular}}
    \label{tab:label_difference}
\end{table}

We \textit{roughly} compute the number of main additive labels adopted in different approaches, containing Pivoting \cite{gu2018unpaired}, Song \textit{et al.} \cite{song2019unpaired}, Guo \textit{et al.} \cite{guo2019mscap}, Feng \textit{et al.} \cite{feng2019unsupervised}, Laina \textit{et al.} \cite{laina2019towards}, Cao \textit{et al.} \cite{cao2020interactions}, Gu \textit{et al.} \cite{gu2019unpaired}, SCS \cite{ben2021unpaired}, and WS-UIC, shown in Table \ref{tab:label_difference}. There are mainly four types of additive labels adopted in the related approaches, including BBox labels, image-caption pairs, relationship-triplet labels, and image-level labels. From Table \ref{tab:label_difference}, we can clearly see that all of the previous approaches depended on enormous of the former three costly labels. In sharp contrast, WS-UIC relies on only cheap image-level labels. For example, Gu \textit{et al.} \cite{gu2019unpaired} utilized around 3.84 million BBox labels and 2.35 million relationship-triplet labels, and SCS \cite{ben2021unpaired} employed about 3.84 million BBox labels.

To compare the labeling prices of all these related approaches, we roughly compute them according to the \textit{Built-in workflow with Amazon Mechanical Turk} \footnote{\url{https://aws.amazon.com/sagemaker/groundtruth/pricing/}}. The labeling price per object per review instance is clearly marked, including \$0.012 for each image-level label in the image classification task, \$0.036 for each BBox label, etc. Since each relationship-triplet label consists of three BBox labels, \textit{i.e.}, (subject, relationship, object), so we regard its unit price as $\$0.036*3 = \$0.108$ although we know the actual price is higher than \$0.108. And Amazon Mechanical Turk recommends using multiple labelers per object to improve label accuracy. Thus, we assume using three labelers and the labeling cost of each approach is shown in the last column of Table \ref{tab:label_difference}. For example, the labeling cost for Gu \textit{et al.} \cite{gu2019unpaired} is $(3.84M * \$0.036 + 2.35M * \$1.08) * 3 \approx \$1176.1k$. From the Table, we can see that the labeling price of our work is significantly less than other works. For instance, the labeling price of Gu \textit{et al.} \cite{gu2019unpaired} is about 30 times that of ours, and the labeling price of SCS \cite{ben2021unpaired} is almost 10 times that of ours. To summarize, WS-UIC is cost-effective and has the best generalization ability among all these approaches.

%The price is not accurate since we ignore many complicated factors, such as the price of Amazon SageMaker training and filter out high-quality labels. Besides the price, the labeling cost also includes the labeling time, the number of available annotators, etc. Obviously, the labeling cost of other works will be even much higher than ours when we consider more factors. 

\subsection{Ablation Studies}

\noindent \textbf{Component Analysis. } To help readers have a deeper understanding of our model, we conduct extensive experiments for the component analysis on the test split of the COCO dataset. As shown in Table \ref{tab:ablation_study}, we compare with the following models: 

\emph{\textbf{Obj:}} The WS-OR module alone is applied in this experiment, which is employed to recognize the object instances. Then, these object instances provide accurate object-word alignments with images for the UIC model. As a result, the approach leads to the worst results.

\emph{\textbf{Obj+Rel:}} Both the WS-OR model and the WS-RR model are used in the experiment. The relationship classifier reasons the association information between different objects. Then, these relationship concepts can be implemented to provide more word alignments with images. The boosted experimental performance shows that the relationship concepts can be implemented to benefit the model training.

\emph{\textbf{Obj+UnO:}} Besides the WS-OR, the unrecognized object loss is also leveraged to train the UIC. If one object concept is contained in the captions generated by the UIC model, but not recognized by the WS-OR, a loss will be given to the UIC model. The experimental performance is improved compared with approach \textit{\textbf{Obj}}, which verifies the usefulness of the unrecognized object loss.

\emph{\textbf{Obj+UnO+Rel:}} All the modules of WS-UIC are adopted in this approach, including the WS-OR model, the WS-RR model, and the unrecognized object loss. The experimental performance outperforms all aforementioned approaches under the same unpaired dataset settings, which illustrates the effectiveness of the combination of these models.

\emph{\textbf{WS-UIC:}} Besides the WS-OR model, the WS-RR model, and the unrecognized object loss, we filtered out the captions with at least one object concept as the pseudo captions to train a fully-supervised image captioning model with the cross-entropy loss. The experimental performance surpasses all other approaches, which validates the promising ability of the proposed WS-UIC.

The performance of different UIC approaches exhibits that the proposed WS-OR and WS-RR have the ability to learn the alignments between the words and images. In addition, the experiments related to the unrecognized object loss also illustrate the capability of avoiding the captions containing wrong object concepts.

\noindent \textbf{Effect of the GNN-BR in WS-RR. } 
Compared to the standard GNN \cite{ahn2019weakly}, we add the batch normalization layer and the residual block, therefore, we can get the GNN-BR. As shown in Table \ref{tab:GNN-BR}, our newly proposed GNN-BR beats the GNN model on all the five evaluation metrics. For example, we get 19.7 and 62.1 on B4 and C metrics, respectively, while the GNN model only achieves 18.6 and 59.5. These experiments all validated the effectiveness of our improved GNN model.

\begin{table}[!t]\small
    \centering
    \caption{Ablation study of our WS-UIC with different architectures of WS-RR (\textit{i.e.}, GNN and our GNN-BR).}
    \setlength{\tabcolsep}{1mm}{
    \begin{tabular}{ccccccc}
    \hline
   \multirow{1}{*}{\makecell{}}  & B4 & M & R & C  &  S\\ \hline
     \multirow{1}{*}{\makecell{\textit{GNN}} }
   & 18.6&19.1&43.9&59.5&12.7\\ \hline
 \multirow{1}{*}{\makecell{ \textit{GNN-BR}} }  
 &\textbf{19.7}  &\textbf{19.6}&\textbf{44.5}&\textbf{62.1}& \textbf{13.0} \\ \hline
  %     WS-UIC& \XSolidBrush  &&&&&\\   \hline
    \end{tabular}}
    \label{tab:GNN-BR}
\end{table}

\begin{figure}[!tbp]
    \centering
    \includegraphics[width=8cm, height=5cm]{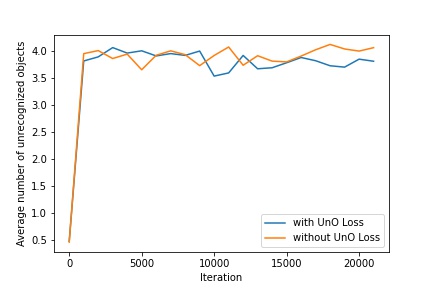}
    \caption{The average number of unrecognized objects in the generated captions during the training process.}
\label{fig:Qua_result_num_unobject}
\end{figure}

\noindent \textbf{Effect of Unrecognized Object Loss.} 
To illustrate the usefulness of the unrecognized object loss, the number of unrecognized objects in the training process are computed and exhibited in Fig. \ref{fig:Qua_result_num_unobject}. We can find that the number of unrecognized object categories is decreasing slightly when the loss is utilized in the model, containing about 3.84 unrecognized object concepts on average. The number of unrecognized object concepts is not decreasing when the UIC model without the loss, containing about 3.93 unrecognized object concepts on average. The number of unrecognized object concepts is so high even we use the unrecognized object loss. It is because that we only filter out concepts with much high confident scores and ignore the concepts with low confident scores. All in all, our proposed unrecognized object loss contributes to the UIC model and generates fewer unrecognized object classes.

\subsection{Qualitative Results}
\begin{figure*}[!tbp]
    \centering
    \includegraphics[width=17cm, height=10cm]{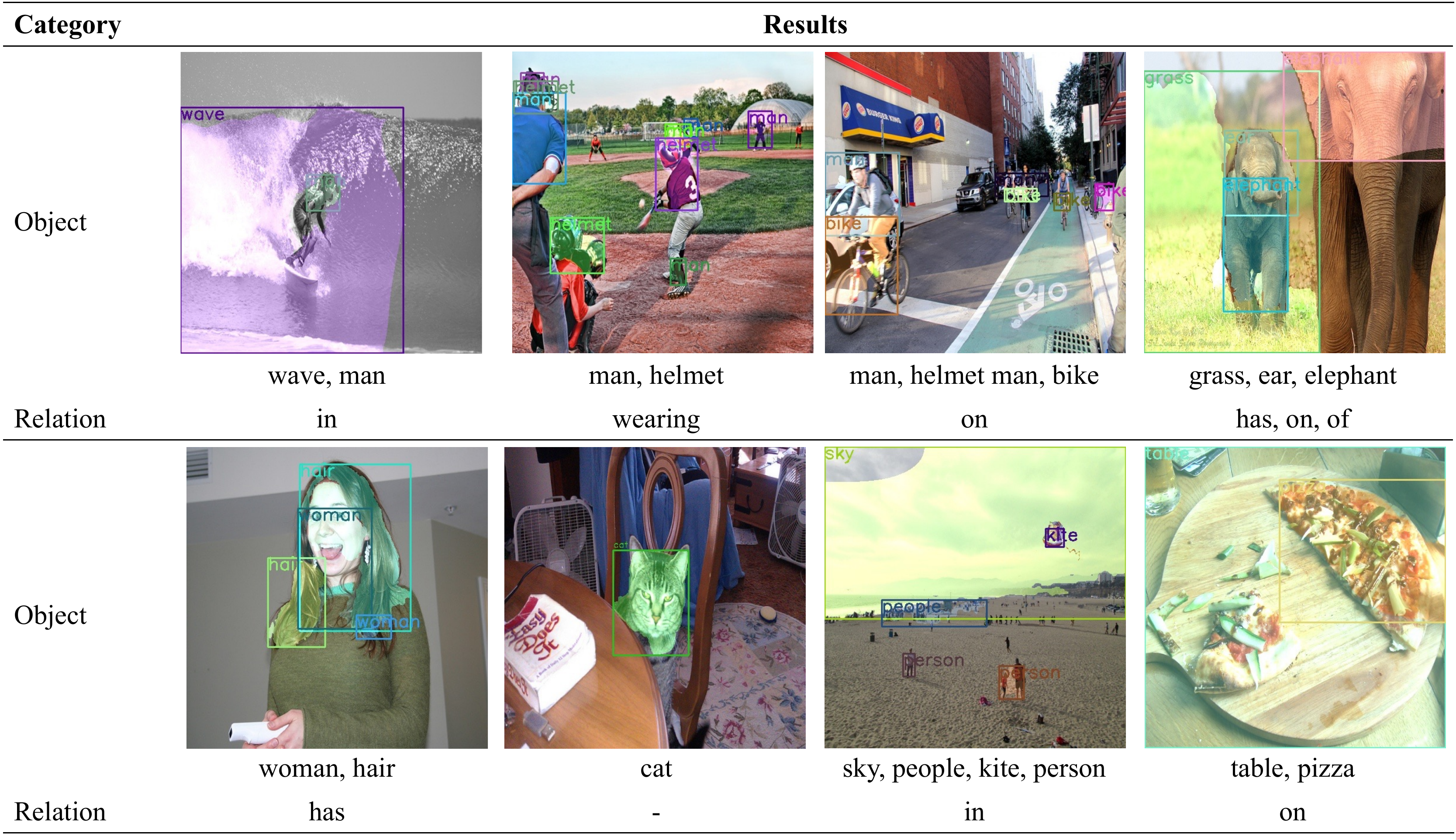}
    \caption{The visualization of the object information and relation information, which is generated by the WS-OR and WS-RR, respectively.}
\label{fig:Qua_result_v}
\end{figure*}
\begin{figure*}[!tbp]
    \centering
    \includegraphics[width=17cm, height=9cm]{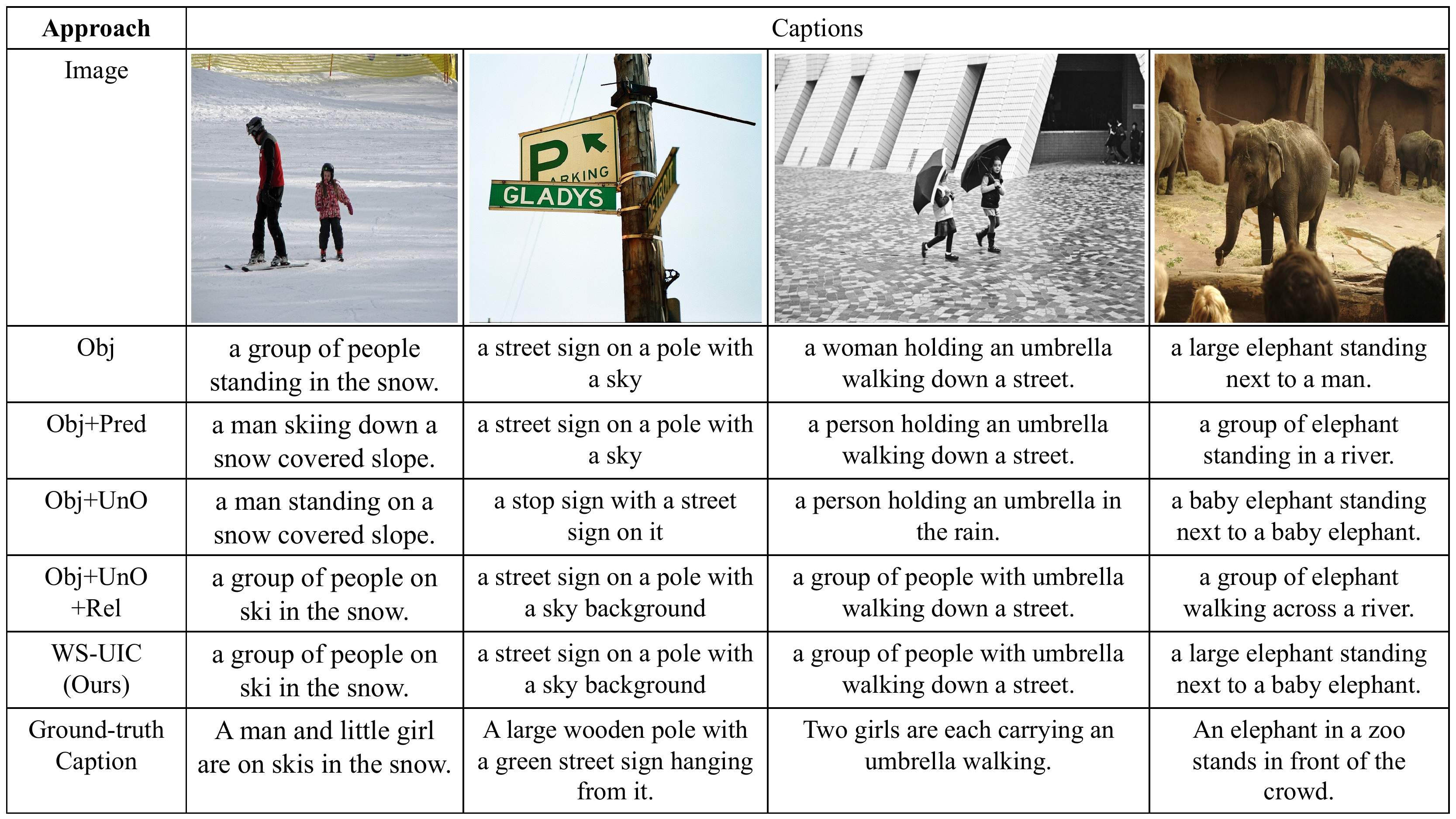}    \caption{Qualitative captions of different approaches with ground-truth captions.  The ``Obj" line exhibits the captions generated by UIC with object information. The ``Obj+Rel" line elaborates the captions generated by UIC with both object information and relationship information. The ``Obj+UnO" line elaborates the captions generated by UIC with both object information and the unrecognized object loss. The ``Obj+UnO+Rel" line represents the captions generated by UIC with object information, relationship information, and the unrecognized object loss. The ``WS-UIC" line displays the captions generated after the pseudo caption training by a fully-supervised image caption model. The ground-truth captions are shown in the last line for comparison.}
\label{fig:Qua_result}
\end{figure*}

\noindent \textbf{Visualization of WS Visual Concept Recognition. } 
In this section, to give a qualitative evaluation on the WS visual concept recognition, we provide some visualization of object and relation information predicted by our WS-OR and WS-RR module. As shown in Fig. \ref{fig:Qua_result_v}, we can find that the WS-OR is able to differentiate the salient and main object categories with locations. These examples also illustrate the strong ability of WS-RR for recognizing the relations between two objects. Let us take the first image as an example, the object category ``wave" and ``man" are clearly and correctly identified by the WS-OR. For the relationship between the ``wave" and ``man", the concept ``in" is given by the WS-RR model considering the location information of the two objects, which demonstrates the promising differentiation capability of WS visual concept recognition. All these examples fully demonstrate the effectiveness of our model.

\noindent \textbf{Qualitative Results of WS-UIC.} 
As shown in Fig. \ref{fig:Qua_result}, we provide some representative captions generated with multiple models, including Obj, Obj+Rel, Obj+UnO, Obj+UnO+Rel, and WS-UIC. The Obj means UIC with object information. %The Obj+Rel represents UIC with both object information and relationship information. The Obj+UnO means UIC with both object information and unrecognized object loss. The Obj+UnO+Rel is UIC with object information, unrecognized object loss, and the relationship information. Besides all the proposed modules, the WS-UIC is further trained a fully-supervised image-caption model by using the generated captions as the pseudo captions. We also introduce the ground-truth captions to help readers better understand the image. 

From these results, we can find that these UIC models can generate reasonable captions by using the objects and the relationships recognized through the proposed WS visual concept recognition scheme. Take the first image as an example, the correct term ``people" and ``snow" are presented in the approach ``Obj". The caption of the approach ``Obj + Pred" includes ``slope" and ``man" which details the object concepts further. The information ``standing on" exists in the caption of approach ``Obj + UnO", which represents the accurate state of the ``man". The generated captions of ``Obj+UnO+Rel" are more accurate and descriptive than captions of other approaches, which spot the information of ``a group of people", ``on ski", and ``in the snow" of the image. The captions of ``WS-UIC" is the same as ``Obj+UnO+Rel". Take the last image as an example, the caption of the approach ``Obj+UnO+Rel" contains the incorrect ``river", and no incorrect concept is generated by the approach ``WS-UIC". These qualitative results strongly validate the merit of the proposed WS object recognition, the usefulness of the WS relationship recognition, the exclusion ability of the designed unrecognized objects loss, and the helpfulness of the pseudo-caption training.

\section{Limitations and Discussions}

\begin{figure}[!tbp]
    \centering
    \includegraphics[width=7.5cm, height=7.5cm]{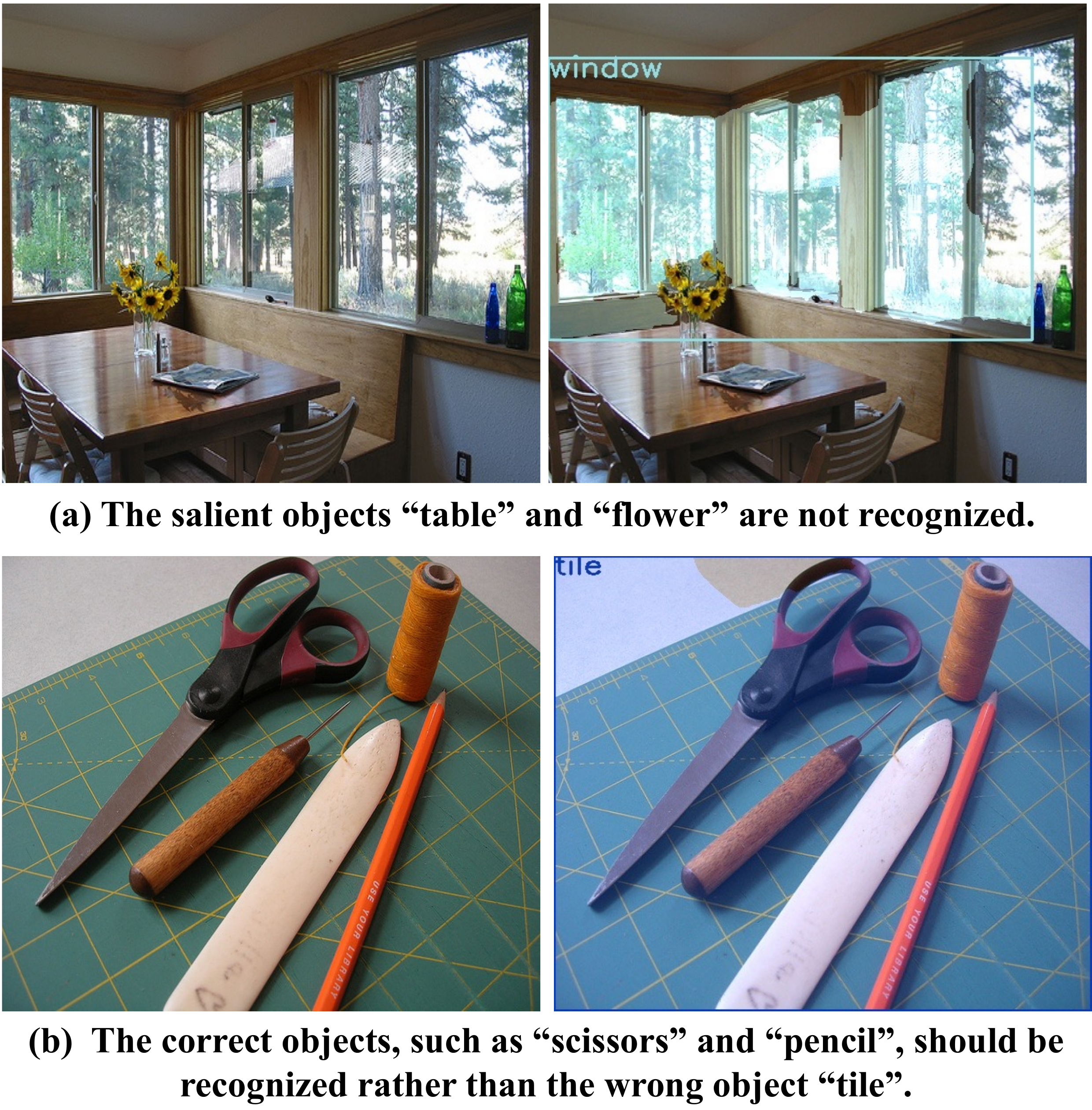}
    \caption{Two failing examples for the WS visual concept recognition.}
\label{fig:failed_example_seg}
\end{figure}

Despite that the unpaired image captioning is achieved by utilizing the proposed model of WS visual concept recognition, some issues are still required to be addressed. The recognized visual concepts can provide paired visual concepts for images, but the WS schemes cannot accurately capture each of the salient concept information. For instance, the ``table" and ``flower" are not recognized in the first image of Fig. \ref{fig:failed_example_seg}. The accuracy of the visual concepts directly influences the performance of unpaired image captioning. A performance gap still exists between WS-UIC and some of the approaches with stronger supervision. It will be a promising direction to utilize more advanced weakly-supervised segmentation models for scene understanding \cite{zhang2021segmenting}.

% Another limitation is that a large number of the visual concepts cannot be recognized in an image. Although a certain number of object and relationship categories are adopted in WS-UIC, there are still enormous categories not involving. Take the second image of Fig. \ref{fig:failed_example_seg} as an example, the object concepts ``pencil", ``ruler", ``scissors", and ``thread" are not included in the category corpus so these objects will not be differentiated correctly. Additionally, the approach cannot differentiate other types of concepts, such as attribute concepts, which are also the important components in a description. If the scheme can be employed to recognize more types of concepts, the performance of unpaired image captioning will be improved.

\begin{figure}[!tbp]
    \centering
    \includegraphics[width=8cm, height=4.5cm]{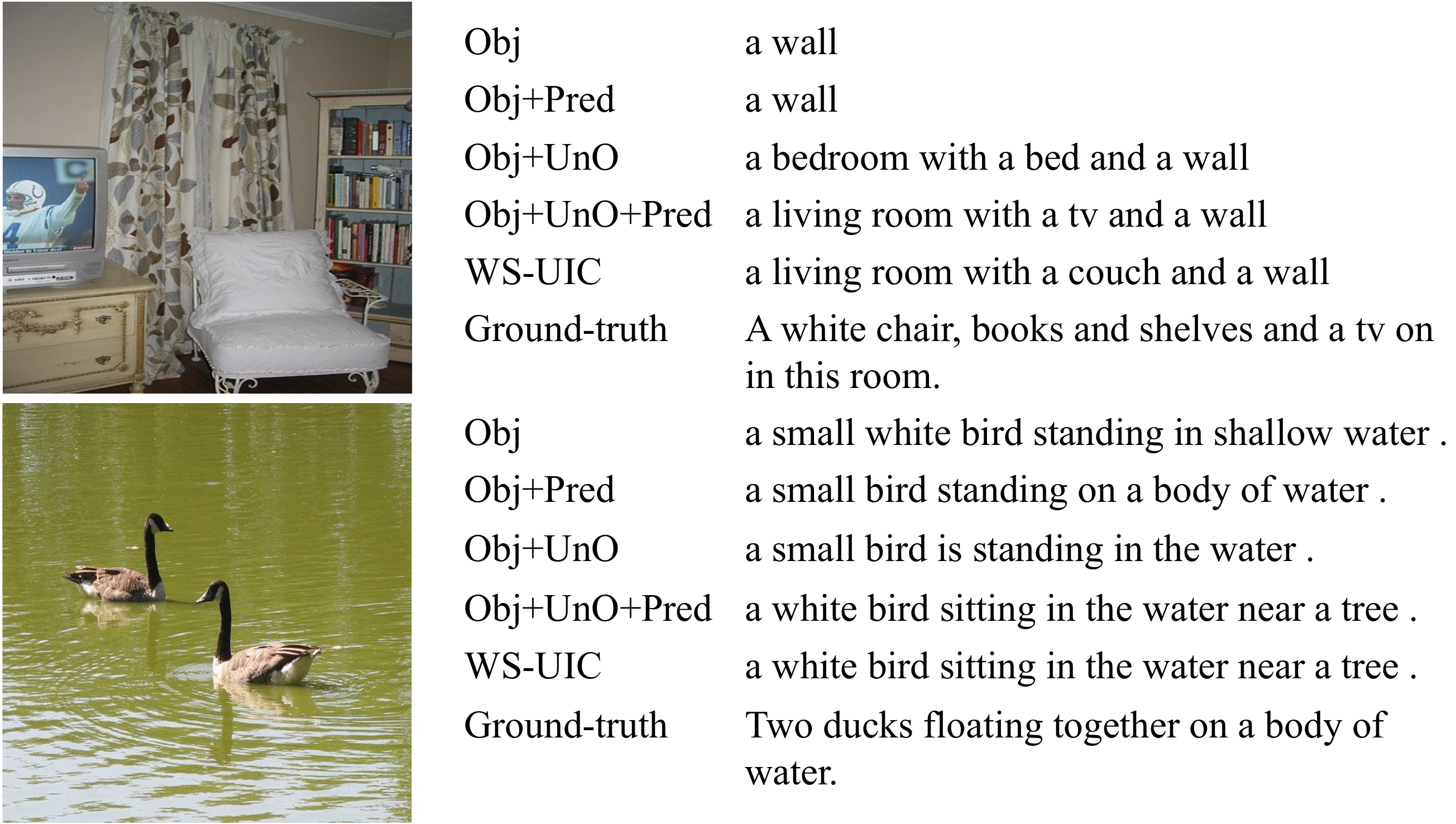}
    \caption{Two failing examples for the WS-UIC.}
\label{fig:failed_example_captions}
\end{figure}

The second one is about the quality of the generated descriptions. Given the recognized concepts, the ability of the UIC model plays a vital role in describing images. A better model is that it can generate a correctly structured description with reasonable rhetoric by aligning the image features and these concepts. However, our model needs to be improved at this aspect according to the simple sentences with wrong concepts exhibited in Fig. \ref{fig:failed_example_captions}. In our future works, we will exploit strong network architectures (e.g., the Transformer \cite{vaswani2017attention, liu2021Swin}), for better image caption generation.

\section{Conclusion} % and Future Works
%In this paper, we have presented a novel scheme named WS-UIC to extract the object concepts and relationship concepts using only the image-level labels. In contrast to the conventional UIC approaches, the proposed scheme is able to generate detailed descriptions for an image without relying on the labels generated through costly process. More specifically, the object concepts have been derived by implementing the weakly supervised instance segmentation while the relationship concepts have been generated by building an enhanced graph neural network. It is worth noting that the recognized relationships in an image can be inferred by combining the information of objects as obtained. Extensive experiments have been conducted to demonstrate the effectiveness of WS-UIC by comparing it with the existing approaches that require much stronger supervision.
In this paper, we have presented a novel cost-effective framework for weakly-supervised unpaired image captioning (WS-UIC).
%termed WS-UIC to extract the object concepts and relationship concepts using only the \textit{image-level} labels.
In contrast to the conventional UIC approaches that rely on expensive training labels, the proposed method is able to generate detailed descriptions for an image given only image-level labels.
Besides, an enhanced graph neural network and a novel unrecognized object loss were designed to improve the performance of unpaired image captioning.
%More specifically, the object concepts have been derived by implementing the weakly supervised instance segmentation while the relationship concepts have been generated by building an enhanced graph neural network. It is worth noting that the recognized relationships in an image can be inferred by combining the information of objects as obtained. Moreover, an unrecognized object loss is proposed to avoid the UIC model generating captions with wrong object concepts.
Extensive experiments have been conducted to demonstrate the effectiveness of the proposed WS-UIC scheme by comparing it with several competitive counterparts that require much stronger supervision. The proposed scheme can be further enhanced by incorporating some more sophisticated strategies, such as cross-alignment between the vision domain and the language domain.
%alignment between these two domains to improve the caption generation for UIC.
%In the future, we can add attribute information into the scheme given only image-level attribute labels. Additionally, it is promising to explore visual concepts with only image-level object labels and no relationship/attribute labels are used.
%-------------------------------------------------------------------------
\bibliography{aaai22}
\bibliographystyle{IEEEtran}

\end{document}